\documentclass{article}
\PassOptionsToPackage{numbers,compress}{natbib}
\usepackage[preprint]{neurips_2026}

\usepackage[utf8]{inputenc} %

\usepackage{graphicx}

\usepackage[T1]{fontenc}

\usepackage[hidelinks,colorlinks]{hyperref}       %
\usepackage{url}            %
\usepackage{booktabs}       %
\usepackage{amsfonts}       %
\usepackage{nicefrac}       %
\usepackage{microtype}      %
\usepackage{pgf}

\IfFileExists{headers/config/showoverfull.config}{
	\overfullrule=1cm
}{
}

\usepackage{etoolbox}

\newbool{includeappendix}
\setbool{includeappendix}{true} %
\IfFileExists{headers/config/noappendix.config}{
	\setbool{includeappendix}{false}
}{}

\newif\ifincludeappendixx
\ifbool{includeappendix}{
	\includeappendixxtrue
}{
	\includeappendixxfalse
}

\usepackage{xr} %
\usepackage{filecontents}

\usepackage{xspace}

\newcommand{\eg}{e.g., }
\newcommand{\ie}{i.e., }

\newcommand{\kgwd}{\textsc{KGW-D}\xspace}

\usepackage{acro} %

\usepackage{listings}

\usepackage{textcomp}

\usepackage{xcolor}

\usepackage[scaled=0.8]{beramono}

\definecolor{ckeyword}{HTML}{7F0055}
\definecolor{ccomment}{HTML}{3F7F5F}
\definecolor{cstring}{HTML}{2A0099}

\lstdefinestyle{numbers}{
	numbers=left,
	framexleftmargin=20pt,
	numberstyle=\tiny,
	firstnumber=auto,
	numbersep=1em,
	xleftmargin=2em
}

\lstdefinestyle{layout}{
	frame=none,
	captionpos=b,
}

\lstdefinestyle{comment-style}{
	morecomment=[l]//,
	morecomment=[s]{/*}{*/},
	commentstyle={\color{ccomment}\itshape},
}

\lstdefinestyle{string-style}{
	morestring=[b]",%
	morestring=[b]',%
	stringstyle={\color{cstring}},
	showstringspaces=false,%
}

\lstdefinestyle{keyword-style}{
	keywordstyle={\ttfamily\bfseries},
	morekeywords={
		function,
		constructor,
		int,
		bool,
		return,
		returns,
		uint
	},
	morekeywords = [2]{},
	keywordstyle = [2]{\text},
	sensitive=true,
}

\lstdefinestyle{input-encoding}{
	inputencoding=utf8,
	extendedchars=true,
	literate=
	{ℝ}{$\reals$}1%
	{→}{$\rightarrow$}1%
	{α}{$\alpha$}1%
	{β}{$\beta$}1%
	{λ}{$\lambda$}1%
	{θ}{$\theta$}1%
	{ϕ}{$\phi$}1%
}

\lstdefinestyle{escaping}{
	moredelim={**[is][\color{blue}]{\%}{\%}},
	escapechar=|,
	mathescape=true
}

\lstdefinestyle{default-style}{
	basicstyle=\fontencoding{T1}\ttfamily\footnotesize,
	style=numbers,
	style=layout,
	style=comment-style,
	style=string-style,
	style=keyword-style,
	style=input-encoding,
	style=escaping,
	tabsize=2,
	upquote=true
}

\lstdefinelanguage{BASIC}{
	language=C++,
	style=default-style
}[keywords,comments,strings]%

\makeatletter
\@ifundefined{ificmlsubmission}{\newif\ificmlsubmission}{}
\makeatother

\usepackage[utf8]{inputenc} %
\usepackage[T1]{fontenc}    %
\usepackage{caption}
\usepackage{enumerate}
\usepackage{url}            %
\usepackage{booktabs}       %
\usepackage{amsfonts}       %
\usepackage{nicefrac}       %
\usepackage{microtype}      %
\usepackage{xcolor}         %
\usepackage{tikz,booktabs,multirow,enumitem,bm}
\usepackage{colortbl}
\usepackage{adjustbox}
\usepackage{placeins} %
\definecolor{hlgreen}{RGB}{198,239,206}
\usepackage{tabularx}
\usepackage{subcaption}
\usepackage{wrapfig}
\usepackage{tabulary}
\usepackage{amsmath,amsthm,amsfonts, bbm}
\usepackage{makecell}
\usepackage{rotating}
\usepackage{array}
\usepackage{siunitx}
\usepackage[most]{tcolorbox}
\ificmlsubmission
\else
  \usepackage{algpseudocode}
  \usepackage{algorithm}
\fi
\usepackage{CJKutf8}
\usepackage{pifont}

\usepackage{amsmath,amsfonts,bm}
\usepackage{amsthm}
\usepackage{amssymb}
\usepackage{mathtools}

\def\1{\bm{1}}

\DeclareMathAlphabet{\mathsfit}{\encodingdefault}{\sfdefault}{m}{sl}
\SetMathAlphabet{\mathsfit}{bold}{\encodingdefault}{\sfdefault}{bx}{n}

\newcolumntype{x}[2]{S[table-format=#1.#2,table-auto-round]}
\newcolumntype{y}[2]{>{\small} S[table-format=#1.#2,table-auto-round]}

\definecolor{hyperlinkblue}{HTML}{0000AA}
\hypersetup{citecolor=hyperlinkblue} %

\definecolor{red}{HTML}{FF0000}
\definecolor{drop}{HTML}{2596be}
\definecolor{anon}{HTML}{10a37f}
\definecolor{devil}{HTML}{84cf9d}

\lstdefinestyle{mystyle}{
    escapechar=\#,
    breaklines=true,
    basicstyle=\scriptsize\ttfamily,
    numbers=none,
    language={},
    framextopmargin=0pt,
    framexbottommargin=0pt,
    breakindent=0pt,
    showspaces = false,
    keywordstyle=\bfseries,
    showstringspaces=false,
    columns=fullflexible,
    morekeywords={Style, Consistency, Accuracy, Ethics, Score}
}

\newtcblisting{prompt}[2][]{
    arc=3pt, outer arc=3pt,
    width=.9\linewidth,
    left=1mm,
    top=0mm,
    bottom=0mm,
    title=#2, 
    colback=gray!5!white,
    colframe=black!75!black,
    fonttitle=\bfseries,
    listing only, 
    listing options={style=mystyle},
    breakable,
    #1
}

\newtcblisting{response}[2][]{
    arc=3pt, outer arc=3pt,
    width=.9\linewidth,
    left=1mm,
    top=0mm,
    bottom=0mm,
    title=#2, 
    colback=devil!5!white,
    colframe=devil!50!black,
    fonttitle=\bfseries,
    listing only, 
    listing options={style=mystyle,deletekeywords={Style}},
    breakable,
    #1
}

\newtcblisting{inference}[2][]{
    arc=3pt, outer arc=3pt,
    width=.9\linewidth,
    left=1mm,
    top=0mm,
    bottom=0mm,
    title=#2, 
    colback=red!5!white,
    colframe=red,
    fonttitle=\bfseries,
    listing only, 
    listing options={style=mystyle},
    breakable,
    #1
}

\newtcblisting{anonymize}[2][]{
    arc=3pt, outer arc=3pt,
    width=.9\linewidth,
    left=1mm,
    top=0mm,
    bottom=0mm,
    title=#2, 
    colback=anon!5!white,
    colframe=anon,
    fonttitle=\bfseries,
    listing only, 
    listing options={style=mystyle},
    breakable,
    #1
}

\lstdefinelanguage{QA}{
  morekeywords={Prompt,Base,Watermarked},
  sensitive=false,
}

\lstdefinestyle{qaStyle}{
  language=QA,
  basicstyle=\normalfont\small,      %
  keywordstyle=\bfseries\color{teal},%
  stringstyle=\color{black},         %
  numbers=none,                       %
  showstringspaces=false,            %
  frame=single,                      %
  rulecolor=\color{gray!50},         %
  backgroundcolor=\color{gray!10},   %
  columns=fullflexible,              %
  xleftmargin=1em, xrightmargin=1em, %
  aboveskip=1em, belowskip=1em       %
}

\definecolor{outerbg}{HTML}{F6F6F6}      %
\definecolor{outerframe}{HTML}{D0D0D0}   %

\definecolor{modelAback}{HTML}{DD8452}   %
\definecolor{modelAframe}{HTML}{DD8452}  %

\definecolor{modelBback}{HTML}{4C72B0}   %
\definecolor{modelBframe}{HTML}{4C72B0}  %

\newtcolorbox{promptbox}[1][]{%
  breakable,
  colback=white,
  colframe=black!25,
  fonttitle=\bfseries\small,
  coltitle=black!80,
  title=User~Prompt~{#1},
  boxsep=5pt,
  top=4pt,bottom=4pt,
  arc=1mm,
}

\newtcolorbox{modelbox}[3][]{%
  breakable,
  colback=#2!75,
  colframe=#3!85,
  coltitle=#3!10!black,
  fonttitle=\bfseries\small,   %
  title={#1},
  boxsep=5pt,
  top=4pt,bottom=4pt,
  arc=1mm,
}

\renewcommand{\S}{Sec.~}

\usepackage[capitalize,noabbrev]{cleveref}

\crefformat{section}{\S#2#1#3}

\crefrangeformat{section}{\S#3#1#4\crefrangeconjunction\S#5#2#6}

\crefmultiformat{section}{\S#2#1#3}{\crefpairconjunction\S#2#1#3}{\crefmiddleconjunction\S#2#1#3}{\creflastconjunction\S#2#1#3}

\newcommand{\crefrangeconjunction}{--}

\crefname{listing}{Lst.}{listings}
\crefname{line}{Lin.}{Lin.}
\crefname{appendix}{App.}{App.}

\newcommand{\appref}[1]{%
	\ifbool{includeappendix}{App.~\ref{#1}}{the appendix}%
}
\newcommand{\Appref}[1]{%
	\ifbool{includeappendix}{App.~\ref{#1}}{The appendix}%
}
\newcommand{\apprefrange}[2]{%
	\ifbool{includeappendix}{App.~\ref{#1}--\ref{#2}}{the appendix}%
}

\usepackage{aliascnt}
\theoremstyle{plain}

\newaliascnt{lemma}{theorem}

\aliascntresetthe{lemma}
\crefname{lemma}{Lemma}{Lemmas}
\Crefname{lemma}{Lemma}{Lemmas}

\theoremstyle{definition}

\theoremstyle{remark}

\newcommand{\OurTitle}{Making Open-Source Text LLM Watermarks Durable Against Merging}

\title{\OurTitle{}}

\author{
  Luisa Scharff, Thibaud Gloaguen, Robin Staab, Martin Vechev\\
  ETH Zurich\\
  \texttt{lscharff@ethz.ch}
}

\begin{document}

\maketitle

\begin{abstract}
	Open-source LLMs (OSMs) are reaching near state-of-the-art performance, prompting prior works to trace the text they generate by embedding text watermarking algorithms directly into their weights.
Yet, OSMs are subject to post-training modifications, which has been shown to remove the watermark.
Model merging in particular, a prominent method used for combining expert knowledge and preventing catastrophic forgetting, strongly removes such OSMs watermarks. 
A key question is how to enable OSM watermarks that survive subsequent merging.
In this work, we show for the first time how to design an OSM watermark that is durable against model merging.
We propose Merge-Adversarial Training, an adversarial training algorithm to distill text watermarks into model weights while being robust to subsequent model merging.
Our approach consistently outperforms all baselines (e.g. with SLERP up to +51 percentage points (pp) TPR@1\%FPR with +25 pp on average) while preserving downstream capabilities.
We also for the first time evaluate OSM watermarks against realistic merge scenarios, representing common use-cases such as combining expert capabilities or preventing catastrophic forgetting, and with 3 prominent merging algorithms.
More broadly, our findings suggest that adversarial training is a reliable approach for increasing OSM watermark durability against post-training modifications.

\end{abstract}

\section{Introduction}
\label{sec:introduction}

Large language model (LLM) watermarking is reaching maturity, becoming increasingly integrated into regulatory frameworks~\citep{euai} and consumer-facing products~\citep{synthid}.
Most existing watermarking methods are based on modifying the LLM sampling procedure to imprint a human-invisible but detectable signal in the generated output.
Crucially, these methods are \emph{generation-time} and designed for closed-source models served via an API, and are thus inapplicable to open-source models (OSMs).

\paragraph{Open-source LLM watermarks}
Given the widespread adoption of capable open-source models~\citep{qwen3.5,deepseekai2026deepseekv4,kimiteam2026kimik25visualagentic}, the community has increasingly shifted its focus to OSM watermarks, which directly embed the watermarking behavior into model weights.
Prior work~\citep{gloaguen2025towards} has highlighted a new challenge for OSM watermarks: \emph{durability} under common OSM modifications.
Among these, \emph{model merging}, which combines two or more trained models to join their capabilities, has proven to be surprisingly adverse to OSM watermarks.
This is concerning because model merging is common and cost-effective: merged models routinely rank near the top of leaderboards~\citep{open-llm-leaderboard}, and over $40$ thousand are publicly shared on HuggingFace.
Thus, model merging poses a significant operational challenge: even non-adversarial merges of a released watermarked model may inadvertently remove the watermark.

\ificmlsubmission
\begin{figure*}[t]
\else
\begin{figure}
\fi
    \centering
    \includegraphics[width=\linewidth]{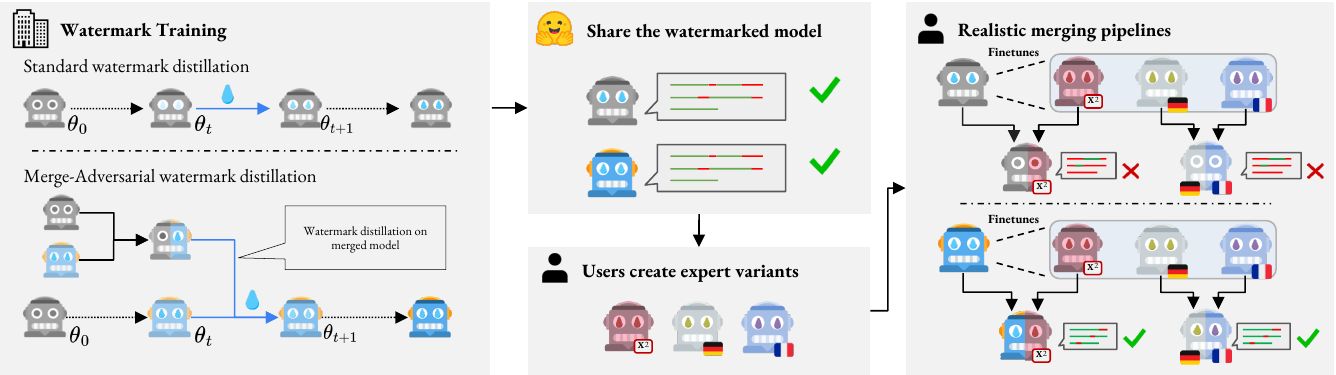}
    \caption{\textbf{Overview of our method and evaluation:} (\emph{Left}) Unlike standard watermark distillation, we train the watermark using an adversarial objective: at each step, we simulate a merge operation and optimize for it to remain watermarked. (\emph{Middle}) After training, both standard distillation and our model are watermarked as long as they are not modified. (\emph{Right}) Yet, users might finetune the model and create complex merged models (\eg to prevent forgetting or to combine capabilities). While standard watermark distillation loses its watermark, our method preserves it consistently.
    }
    \label{fig:accept}
    \vspace{-1em}
\ificmlsubmission
\end{figure*}
\else
\end{figure}
\fi

\paragraph{This work: OSM watermarks durable against merging}
In this work, we introduce the first OSM watermark specifically designed to be durable against model merging.
Specifically, building on the popular watermark-distillation approach~\citep{learnability_wm}, we propose \emph{merge-resistant adversarial training} (MAT), a watermark-distillation objective explicitly designed for robustness to model merging.

MAT builds on an adversarial-training-inspired meta-loss component that directly includes a low-cost merge operation inside the watermark training loop: at each step, we temporarily create a merged model that interpolates the current checkpoint with a reference model, compute the watermark-distillation loss through this merge, and backpropagate gradients only to the current model.
Our experimental evaluation shows that MAT yields a significantly more durable watermark compared to both standard watermark-distillation approaches and other OSM watermarks~\citep{gaussmark} while retaining the same model performance as standard watermark-distillation.

Beyond our method, we introduce a rigorous evaluation pipeline for OSM durability under model merging.
Unlike prior work, which only merges the watermarked model with the unwatermarked base model, we evaluate durability across realistic scenarios, including multiple fine-tuned models, multi-round merging, and three merging algorithms: LINEAR, SLERP, and TIES.
At the same time, our experimental results establish the evaluation of durability under unwatermarked base model merging as an effective worst-case proxy for faster OSM watermarking evaluations.

\paragraph{Our contributions}
We make the following contributions:
\begin{enumerate}[leftmargin=*,itemsep=2pt,topsep=2pt]
    \item We introduce MAT, an adversarial-training-based objective~(\cref{sec:method}) that simulates model merging during training, directly optimizing our watermark to be durable against merging.

    \item We show that MAT consistently improves watermark detectability under realistic merge scenarios and multiple watermarking schemes, significantly outperforming standard watermark distillation while maintaining downstream performance.

    \item We demonstrate that OSM watermarks can be trained in a way that provides additional downstream benefits (\eg durability against merging). 

\end{enumerate}

\section{Background and Related Work}
\label{sec:related_work}

In this section, we review prior work on LLM watermarking, with a particular focus on approaches for open-source models. We further provide the relevant background on model merging.

\paragraph{Generation-time LLM watermarking}
LLM watermarking aims to embed a statistically detectable signal into generated text, enabling later attribution to a particular model or provider.
Most existing schemes operate at generation time: given a private key, they modify the sampling procedure to insert a detectable key-dependent signal into the model output.
The most prominent class is the Red-Green family (KGW)~\citep{kgw1}, where a pseudorandom function partitions the vocabulary into green and red tokens and biases generation toward green tokens.
Other popular schemes include KTH~\citep{kth}, AAR~\citep{aar}, and SynthID~\citep{synthid}.
Unlike Red-Green, these schemes are distortion-free: they preserve the model distribution in expectation over the private key.
However, all methods require modifying the LLM sampling procedure, making them incompatible with the user-controlled open-source setting.

\paragraph{Watermarks for open-source models.}
Watermarking an open-source model requires embedding the signal into the model weights, so that standard inference already produces watermarked text (\ie text with a detectable watermark signal).
This setting has only recently received systematic attention~\citep{gloaguen2025towards}.
Existing approaches broadly fall into gradient-based methods~\citep{learnability_wm, xue2025pro,gloaguen2026llm}, which train the model to internalize a watermark, and weight-editing methods~\citep{gaussmark}, which directly modify model parameters.
Watermark distillation~\citep{learnability_wm} is the most widely used gradient-based approach: a student model is trained to imitate the output distribution of a watermarked teacher, embedding the watermark behavior into its weights.
Meanwhile, weight-editing methods inject the watermark via direct weight modifications requiring no additional training~\citep{gaussmark}.
Crucially, while~\citep{xue2025pro} improves durability against finetuning, no prior work directly improves OSM watermark durability against other common post-training modifications such as model merging.

\paragraph{Robustness, security, and durability}
Prior work on LLM watermark reliability has studied several distinct failure modes.
Text-level robustness asks whether a watermark remains detectable after edits to generated text, such as paraphrasing, translation, deletion, or insertion~\citep{kgw1,kth,kgw2}.
Security work studies adversarial attribution manipulation, including spoofing attacks that generate text falsely detected as watermarked and scrubbing attacks that remove watermarks from generated text~\citep{sadasivan2023can,jovanovic2024watermark,he2024can,pang2024no}.
These directions primarily target generation-time watermarks and text-level adversaries.

Open-source watermarking introduces an additional requirement: \emph{durability} under model modifications.
Because open-source models are routinely finetuned, quantized, and merged after release, useful OSM watermarks should remain detectable after such transformations.
Importantly, recent work evaluating existing OSM watermarks under common post-release modifications finds that current methods remain fragile in these settings~\citep{gloaguen2025towards}: OSM watermarks may survive mild compression-like transformations, but degrade sharply under stronger modifications such as merging.

\paragraph{Model merging}
Model merging constructs a new model by combining the weights of two or more independently trained checkpoints without additional training.
Simple linear interpolation averages model parameters directly, while SLERP~\citep{mergekit} interpolates on the hypersphere.
Task-vector methods such as Task Arithmetic~\citep{taskarithmetic} interpret finetuning updates as directions in weight space that can be added, subtracted, or combined.
TIES~\citep{ties} and DARE-TIES~\citep{dareties} further modify task-vector merging by resolving sign conflicts, pruning updates, or reducing interference between parent models.
These methods are attractive in open-source development because they are inexpensive, require no data or retraining, and can combine specialized finetunes into a single checkpoint.

\section{Our Method}
\label{sec:method}

In this section, we introduce our training algorithm, \emph{Merge-Adversarial Training} (MAT), specifically designed to make OSM watermarks more durable against merging~(\cref{sec:method:training}).
In~\cref{sec:method:scenarios}, we additionally detail all realistic merge scenarios considered for evaluation.

\subsection{Merge-Adversarial Training}
\label{sec:method:training}

\paragraph{Threat model}
We consider a model-provider that releases an OSM watermarked model.
Downstream users can then finetune this model into multiple experts and subsequently merge them freely with one another, additional finetunes, the original base, or community checkpoints.
Importantly, once the watermarked model is released, we assume we have no control over how users use it.

\paragraph{Training setting}
For training, we assume that we have access to an unwatermarked base model.
Our goal is to distill a generation-time watermark into this model as in~\citep{learnability_wm}.
To embed a watermark that is durable against model merging, we adapt standard watermark-distillation using an adversarial training approach.
In particular, we consider model merging as an adversarial action, and regularize our model to be robust against it.
In order to avoid any assumptions on later downstream merging behavior, we instantiate our adversarial merge using the unwatermarked base model.
We show in~\cref{sec:experiments:simplified} that this is, in most cases, a worst-case proxy for watermark durability.

\begin{wrapfigure}[18]{r}{0.6\textwidth}
\vspace{-1.2em}
\begin{minipage}{0.6\textwidth}
\begin{algorithm}[H]

    \caption{Merge-Adversarial Training}
    \label{alg:merge-adversary}
    \begin{algorithmic}[1]
    \Require Frozen base $\theta_0$; watermark transform $w_k(\cdot)$ with top-$k$ gating; dataset $\mathcal{D}$; merge-weight range $[\alpha_{\min}, \alpha_{\max}]$; learning rate $\eta$; training steps $T$

    \State $\theta \gets \theta_0$ \Comment{Initialize trainable copy}

    \For{$s$ from 0 to $T-1$}
        \State $x \gets \text{Sample}(\mathcal{D})$ \Comment{Batch of sequences}
        \State $\tilde{p}(\cdot \mid x_{<t}) \gets w_k\!\left(p_{\theta_0}(\cdot \mid x_{<t})\right)$
        \State $\alpha \gets \text{Sample}\!\left(\mathcal{U}(\alpha_{\min}, \alpha_{\max})\right)$ \Comment{Merge weight}
        \State $\theta_M \gets \alpha\,\theta + (1-\alpha)\,\theta_0$ \Comment{Linear merge}
        \State $\mathcal{L} \gets \displaystyle\sum_{t=1}^{|x|} \mathrm{KL}\!\left(
        \tilde{p}(\cdot \mid x_{<t}) \,\big\|\, p_{\theta_M}(\cdot \mid x_{<t})
        \right)$
        \State $g \gets \nabla_{\theta}\mathcal{L}$
        \State $\theta \gets \theta - \eta g$
    \EndFor \\

    \Return $\theta$

    \end{algorithmic}

\end{algorithm}
\end{minipage}
\vspace{-1.5em}
\end{wrapfigure}

\paragraph{Merge-adversarial training (MAT)}
We present an overview of our method in \cref{alg:merge-adversary}.
Let $\theta_0$ denote the frozen base model and $\theta$ the trainable copy initialised from $\theta_0$ (line 1).
We write $\theta_M$ for a merged model formed by linearly combining $\theta$ with $\theta_0$.
Lastly, $w_k(\cdot)$ corresponds to the generation-time watermarking algorithm, applied to the top-$k$ logits.
It maps the next-token probability distribution to its watermarked counterpart.
At each step we sample a merge weight $\alpha \sim \mathcal{U}(\alpha_{\min}, \alpha_{\max})$ and form the merged parameters $\theta_M = \alpha\,\theta + (1-\alpha)\,\theta_0$ (lines 5-6).
The merged model is trained to mirror the watermarked teacher's next-token distribution using KL divergence (line 7):
\begin{equation}
    \mathcal{L}(\theta) \;=\; \mathbb{E}_{x \sim \mathcal{D}} \sum_{t=1}^{|x|} \mathrm{KL}\!\left( w_k\!\left(p_{\theta_0}(\cdot \mid x_{<t})\right) \,\big\|\, p_{\theta_M}(\cdot \mid x_{<t}) \right),
\end{equation}
where $p_{\theta_0}(\cdot \mid x_{<t})$ and $p_{\theta_M}(\cdot \mid x_{<t})$ correspond to the frozen base $\theta_0$ and the merged $\theta_M$, respectively.
We then backpropagate this loss back to the student model $\theta$ (lines 8-9) to complete our training loop.

\paragraph{Watermark-algorithm $w_k$}
Like standard watermark-distillation, our method is agnostic to the underlying generation-time algorithm.
In \cref{sec:experiments}, we evaluate it extensively on KGW, with additional experiments on the AAR and KTH families.
Unlike \citet{learnability_wm}, however, we find in \appref{sec:ablations:top-k} that watermark distillation should be performed only on the top-$k$ logits to improve model performance.

\paragraph{Choice of merge adversary}
Even though we only explicitely train against linear interpolation with $\theta_0$, we find in \cref{sec:experiments} that the resulting watermark survives non-linear merges as well, suggesting that robustness learned against the simpler merges consistently transfers to more complex methods.
We ablate the merge-weight range $[\alpha_{\min}, \alpha_{\max}]$, the top-$k$ gating, and the watermark strength $\delta$ in \appref{sec:ablations}, and report the full training hyperparameters in \appref{app:wm-distill}.

\subsection{Merge Scenarios and Evaluation Pipeline}
\label{sec:method:scenarios}

Next, we detail all merge scenarios we consider for evaluation, also illustrated in \cref{fig:merge-pipeline}.
We first introduce notation for the parent models involved in each merge, then list the configurations themselves, and lastly describe the resulting evaluation pipeline.

\ificmlsubmission
\begin{figure*}[t]
\else
\begin{figure}[t]
\fi
\centering
\includegraphics[width=\textwidth]{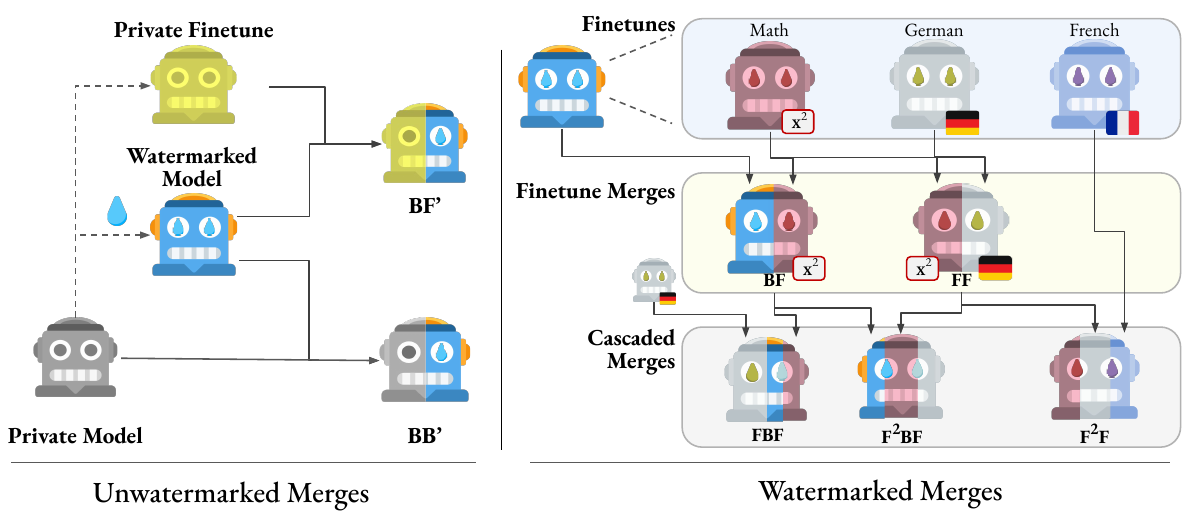}
\caption{\textbf{Merge evaluation pipeline:}
The provider releases a watermarked Instruct base ($B$), and produces watermarked finetunes ($F$) via SFT. Downstream users merge any combination of these models, yielding the scenarios $FF$, $BF$, $F^{2}F$, $FBF$, $F^{2}BF$, $BB'$, and $BF'$ that we evaluate.}
\label{fig:merge-pipeline}
\ificmlsubmission
\end{figure*}
\else
\end{figure}
\fi

Following \cref{sec:method:training}, let $B$ (the watermarked base) correspond to the trained and watermarked $\theta$ released by the provider.
Let $B'$ be the unwatermarked base $\theta_0$. 
We further use $F$ for a finetune of $B$, and $F'$ for an unwatermarked finetune (from the provider).

We group the merges into two families, each corresponding to a real-world deployment scenario.
The \emph{watermarked} family (\cref{fig:merge-pipeline} right) covers four configurations in which every parent is derived from $B$:
(i) the two-expert merge $FF$ (German FT $\otimes$ Math FT), corresponding to a user who has trained multiple domain experts and combines them into a single multi-skilled model;
(ii) the base-expert merge $BF$ ($B$ $\otimes$ Math FT), the standard recipe for mitigating catastrophic forgetting on the FT-target domain by merging a finetune back with the watermarked base;
(iii) the multi-stage cascades $F^{2}BF$ and $FBF$, simulating more complex post-training pipelines such as the recursive expert-and-base merges (\eg used to build \textsc{BgGPT}~\citep{bggpt}), where intermediate $FF$ or $BF$ checkpoints are themselves merged again with a third parent.
We additionally consider $F^{2}F$ (German FT $\otimes$ Math FT $\otimes$ French FT), as an example of a multi-stage merge of individual finetunes (\cref{sec:experiments:realistic}).
For every cascaded merge, the inner merge is fixed at $\alpha = 0.5$ (\eg in $F^{2}BF$ both the inner $FF$ and the inner $BF$ are constructed at $\alpha = 0.5$) while we sweep the outermost $\alpha \in [0.1, 0.9]$ using $0.1$ steps.

The \emph{unwatermarked} family (\cref{fig:merge-pipeline} left) covers the case in which the watermarked base $B$ is merged with a parent that does not carry the watermark: $BF'$ ($B$ $\otimes$ Private Instruct Model Finetune) and $BB'$ ($B \otimes$ Private Instruct Model).
While not reflective of typical deployment scenarios, we find that $BB'$ serves as an empirical lower bound on detection for the harder all-watermarked cascades at matched effective dose (\cref{sec:experiments:simplified}), making it a cheap single-merge proxy for evaluating these configurations.
We note that MAT only optimizes against the linear $BB'$ during training (\cref{sec:method:training}), yet still consistently improves durability across all merge scenarios, including the non-linear ones.

\section{Experimental Evaluation}
\label{sec:experiments}

\definecolor{cgreen}{HTML}{C6EFCE}
\definecolor{corange}{HTML}{FFEB9C}
\definecolor{cred}{HTML}{FFC7CE}

In this section, we show that MAT preserves the performance of standard watermark distillation while being significantly more durable against realistic merges~(\cref{sec:experiments:realistic}).
Following prior work, we also evaluate durability against unwatermarked-parent merges~(\cref{sec:experiments:simplified}) and find that they serve as an effective worst-case proxy for merge-durability evaluation.
In~\cref{sec:experiments:generalization}, we ablate across watermarking algorithms, model architectures, and other OSM watermarks.
We defer additional results to \cref{app:supplementary}.

\paragraph{Experimental setup}
We use \textsc{Llama-3.1-8B-Instruct}, deferring additional results on \textsc{Qwen-2.5-3B-Instruct} to \cref{sec:experiments:generalization}.
We compare MAT against standard watermark distillation (\textsc{KGW-D}) using the same hyperparameters ($\delta=2.3$, $\gamma=0.25$).
Each base model is distilled on a multi-domain mixture spanning English (\textsc{Alpaca-GPT4}~\citep{alpacagpt4}, \textsc{OpenWebText}~\citep{openwebtext}), German (\textsc{Alpaca-GPT4-DE}~\citep{alpacagpt4de}, \textsc{FineWeb-2}~\citep{fineweb2}), math (\textsc{MetaMathQA}~\citep{metamathqa}), and code (\textsc{CodeAlpaca}~\citep{codealpaca}), ensuring watermark learning across domains~\citep{gloaguen2026llm}.
We then derive finetuned versions of each watermarked base using domain-specific datasets: \textsc{NuminaMath-CoT}~\citep{numinamath} for math, \textsc{Evol-Instruct-DE}~\citep{evol-de} for German, and a 58k mix of \textsc{French-Alpaca}~\citep{alpaca-fr} with the Lucie corpus~\citep{lucie} for French.

To measure watermark detectability, we report true positive rate at 1\% false positive rate (TPR@1\%) averaged over $500$ English prompts from C4~\citep{c4}.
Additionally, we measure TPR@1\% on German prompts from FineWeb-2 \texttt{deu\_Latn}~\citep{fineweb2} and math prompts from GSM8K~\citep{gsm8k}, ensuring the watermark persists in finetuned-model domains.
For quality, we report perplexity under \textsc{Llama-2-13B} and \texttt{seq-rep-3}, \ie the fraction of repeated 3-grams in the continuation.
Lower perplexity reflects more natural generations under an external reference judge, while \texttt{seq-rep-3} captures degenerate repetition that perplexity alone can mask.
We further benchmark downstream capabilities with the evaluation harness from~\citet{eval-harness} on ARC \citep{arc}, MMLU \citep{mmlu}, HellaSwag \citep{hellaswag}, GSM8K \citep{gsm8k}, and MATH \citep{math}.
We defer the full training hyperparameters and data compositions to \cref{app:wm-distill}.

\paragraph{Merge scenarios and evaluation}
We follow the threat model from \cref{sec:method:training}: we act as the provider of a watermarked base $B$ that users subsequently merge.
We organize results into two categories.
First, \emph{all-watermarked} merges (\cref{sec:experiments:realistic}), where every parent derives from our watermarked release; here, our method consistently improves merge robustness over the \kgwd baseline.
Second, we test \emph{unwatermarked-parent} merges (\cref{sec:experiments:simplified}), where $B$ is merged with either an external community finetune or the original unwatermarked Instruct base.
In particular, we compare $BB'$ directly against cascaded $FBF$ results at matched effective $\alpha$, and additionally evaluate $BF'$ as a real-world deployment scenario where the watermarked base meets an externally trained community finetune.

\subsection{Durability Against Realistic Merge Scenarios}
\label{sec:experiments:realistic}

\ificmlsubmission
\begin{table}[t]
\else
\begin{wraptable}{r}{0.45\textwidth}
\vspace{-1.3em}
\fi
\centering
\caption{\textbf{Watermark evaluation pre-merge:} TPR@1 and generation quality (PPL; lower is better) of the generation-time KGW (\textsc{KGW}), watermark-distillation (\textsc{KGW-D}), and ours.}
\label{tab:standalone-summary}
\setlength{\tabcolsep}{3pt}
\begin{adjustbox}{width=\linewidth}
\begin{tabular}{l ccc ccc}
\toprule
& \multicolumn{3}{c}{\textbf{TPR@1}} & \multicolumn{3}{c}{\textbf{PPL}} \\
\cmidrule(lr){2-4} \cmidrule(lr){5-7}
\textbf{Variant} & EN & DE & Math & EN & DE & FR \\
\midrule
\textsc{KGW}   & 100.0 & 100.0 & 100.0 & 5.65   & 5.68   &  5.59   \\
\midrule
\textsc{KGW-D} &  98.7 &  98.8 &  95.8 & 5.39 & 5.66 & 5.93 \\
Ours           &  98.0 &  99.0 &  96.4 & 5.69 & 5.59 & 5.46 \\
\bottomrule
\end{tabular}
\end{adjustbox}
\ificmlsubmission
\end{table}
\else
\vspace{-1.5em}
\end{wraptable}
\fi

Here, we evaluate watermark durability under a range of realistic scenarios (illustrated in~\cref{fig:merge-pipeline}).
We first consider two merge configurations widely used in practice: combining two domain-expert finetunes ($FF$) and merging the watermarked base with a single domain finetune to mitigate catastrophic forgetting ($BF$).
We then evaluate three cascaded merges, which stress-test our watermark durability against more complex post-training pipelines.

\paragraph{Watermark detectability-quality trade-off prior to merging}
To ensure a fair comparison, in \cref{tab:standalone-summary}, we verify that, \emph{prior to merging}, our method has a similar detectability-quality trade-off to \kgwd.
This means that our method's improvement in durability against merging is not induced by a stronger watermark.
It also shows that our method does not hurt model performance prior to merging compared to \kgwd{}, both when measuring perplexity~(\cref{tab:standalone-summary}) and benchmark accuracies~(\appref{app:benchmarks:baseline}).

\ificmlsubmission
\begin{figure*}[t]
\else
\begin{figure}[t]
\fi
    \centering
    \includegraphics[width=\textwidth]{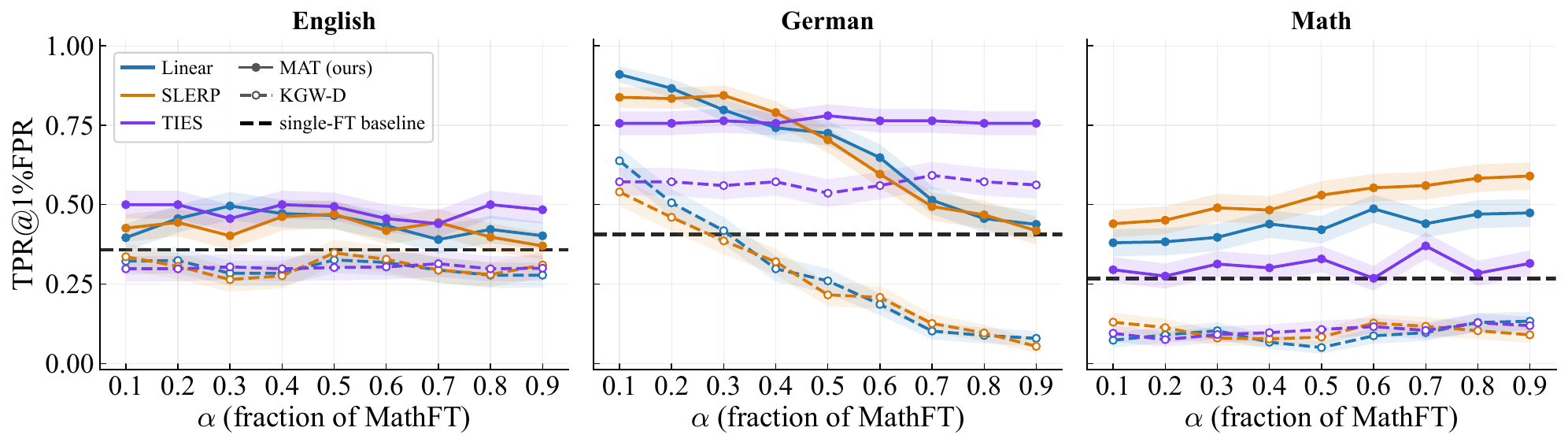}
    \caption{\textbf{Watermark detectability for different merge ratios of experts (FF):} We compare watermark detectability (TPR@1) when merging an $\alpha$ math-finetuned model with a $(1-\alpha)$ German-finetuned model using three different merging algorithms. We average results over $500$ samples with English, German, and math prompts, respectively. The dashed line corresponds to the single-finetuning baseline, \ie the maximum TPR@1 of the finetuned models for the given domain.}
    \label{fig:m1-recovery}
\ificmlsubmission
\end{figure*}
\else
\end{figure}
\fi

\ificmlsubmission
\begin{figure*}[t]
\else
\begin{figure}[t]
\fi
    \centering
    \includegraphics[width=\textwidth]{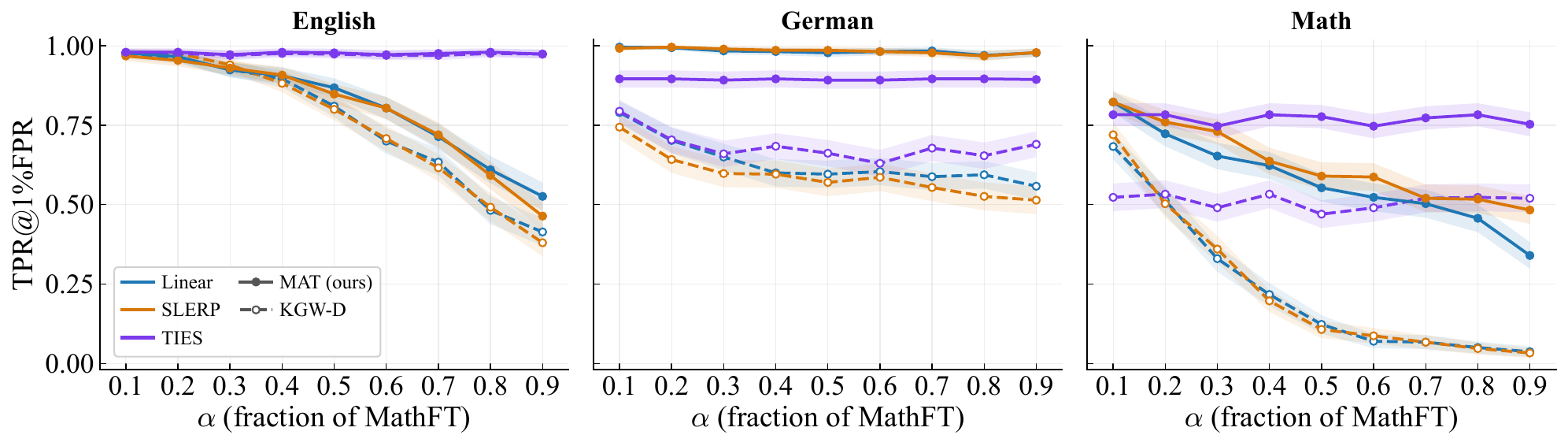}
    \caption{\textbf{Watermark detectability for different merge ratios against catastrophic forgetting (BF):} We compare watermark detectability (TPR@1) when merging an $\alpha$-weighted math-finetuned model with a $(1-\alpha)$-weighted base (watermarked) model using three different merging algorithms. We average results over $500$ samples with English, German, and math prompts, respectively.}
    \label{fig:m2-recovery-main}
\ificmlsubmission
\end{figure*}
\else
\end{figure}
\fi

\paragraph{Our method is durable against finetune merges}
To measure durability against finetune merges (\cref{fig:merge-pipeline}, second row), we compare the merged-model TPR@1\% to the highest parent TPR@1\% on a given domain (called the \emph{single-finetuning baseline}).
Intuitively, a non-durable watermark should yield merged-model TPR@1\% no higher than this baseline.
We also measure watermark detectability on the \emph{expert domains} (\ie German for the German finetune and math for the math finetune), because, as we show in \cref{tab:standalone-detection}, these are the domains where the watermark of the finetuned experts is the weakest (\eg on average, TPR@1\% on the expert domain drops from ${\sim}98\%$ pre-finetuning to ${\sim}15\text{--}25\%$ post-finetuning, while outside this domain detection only drops to ${\sim}28\text{--}40\%$).

In \cref{fig:m1-recovery}, we compare our method with \kgwd in the expert-knowledge combination (FF) scenario.
Across English and the expert domains (German and math), our method systematically outperforms \kgwd.
More importantly, unlike \kgwd, with our method the merged model always has a higher TPR than the single-finetuning baseline (dashed line in \cref{fig:m1-recovery}), which means that our method effectively improves durability against merging by successfully recovering the watermark degradation caused by finetuning.
At the same time, we find in \appref{app:ff-bf-post-merge-benchmarks} that this improvement to durability does not prevent the merged model from successfully combining the expert knowledge.

For the $BF$ merge, corresponding to anti-catastrophic forgetting merges, where one parent is the watermarked base $B$ itself, the watermark is, by construction, better preserved than under $FF$, since one half of the merged weights still encodes the full distillation signal.
Nonetheless, \cref{fig:m2-recovery-main} shows that our method systematically outperforms \kgwd, especially in the expert domain where finetuning has weakened the watermark.
This confirms that our method is robust to finetune merges.

\ificmlsubmission
\begin{figure*}[t]
\else
\begin{figure}[!t]
\fi
\centering
\includegraphics[width=\textwidth]{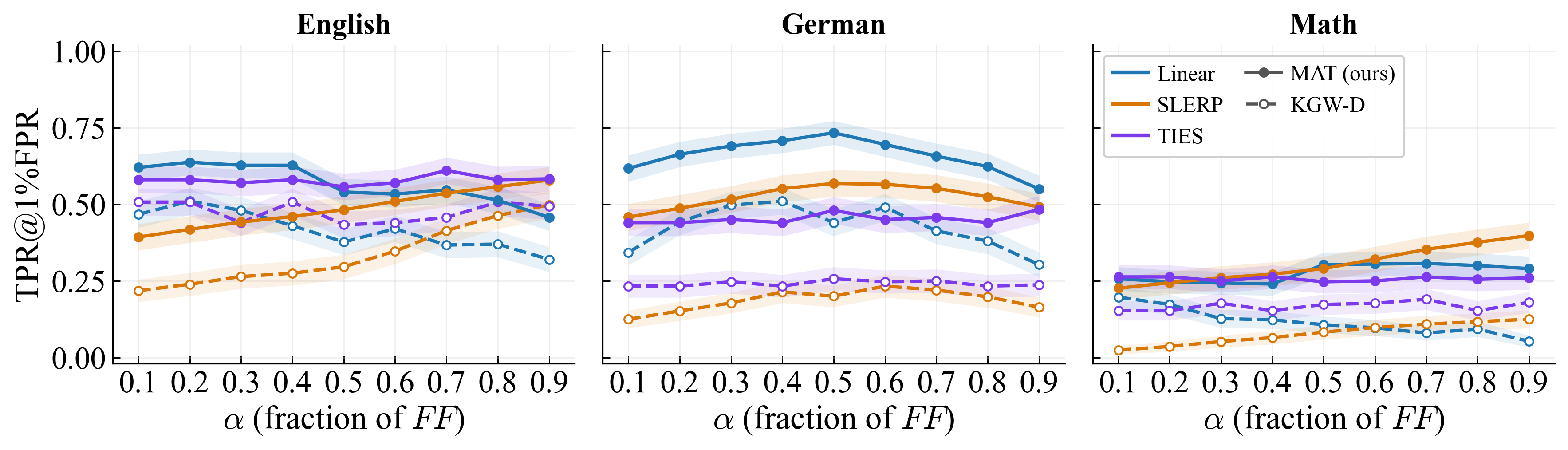}
\caption{\textbf{Watermark detectability with cascaded merges:} We compare watermark detectability (TPR@1) when merging an $\alpha$-weighted $FF$-merged model with a $(1-\alpha)$-weighted French finetuned model using SLERP merging. We average results over $500$ samples with English, German, and math prompts, respectively. The dashed line corresponds to the single-finetuning baseline, \ie the maximum TPR@1 of the $FF$-merged and finetuned models for the given domain.}
\label{fig:french-merge}
\ificmlsubmission
\end{figure*}
\else
\end{figure}
\fi

\paragraph{Our method remains durable against cascaded merges}
We next evaluate whether durability against finetuned merges extends to cascaded merges~(\cref{fig:merge-pipeline}, last row).
Specifically, we consider an $F^{2}F$ configuration: the $FF$ merge above (math finetune merged with German finetune) is merged again with a French finetune.
\cref{fig:french-merge} shows that MAT remains durable, significantly outperforming \kgwd across all evaluated domains and exceeding the single-finetuning baseline. Full per-method results for the remaining cascaded merges ($FBF$, $F^{2}BF$) are in \appref{app:all-watermarked}.

Overall, our results show that our method is durable against all tested merging algorithms and realistic scenarios, unlike prior work.
Importantly, this comes at no cost to model performance before merging.

\subsection{Durability Against Unwatermarked-Parent Merges}
\label{sec:experiments:simplified}

Following prior work~\citep{gloaguen2025towards}, we consider merge scenarios where one parent is unwatermarked.
We verify that these merges are effective proxies for realistic merge scenarios, enabling efficient evaluation of OSM watermark durability under model merging.
\appref{app:unwm-parent-merges} further shows that our watermark is durable across unwatermarked-parent scenarios, including $BB'$ merges with the unwatermarked base and $BF'$ merges with finetunes such as FuseChat~\citep{fusechat}, OpenMath~\citep{openmath}, or Tulu~\citep{tulu}.

\ificmlsubmission
\begin{figure*}[t]
\else
\begin{figure}[t]
\fi
    \centering
    \includegraphics[width=\textwidth]{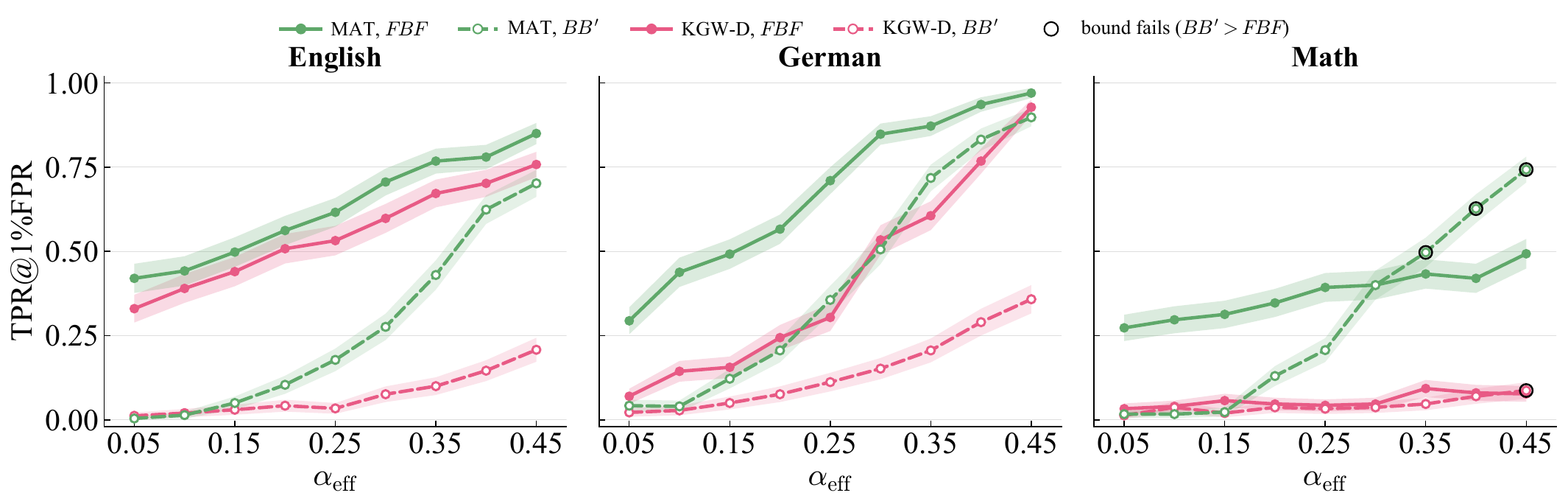}
    \caption{\textbf{Watermark detectability for $BB'$ and $FBF$ with effective merge ratios:} We compare watermark detectability (TPR@1\%) between the unwatermarked merge $BB'$ and $FBF$ at matched effective merge ratio $\alpha_{\text{eff}}$. We highlight points where $BB'$ has higher TPR@1\% than $FBF$.
    }
    \label{fig:lower-bound}
\ificmlsubmission
\end{figure*}
\else
\end{figure}
\fi

\paragraph{Unwatermarked-parent merges as a proxy for realistic merges}
We compare durability against $BB'$ with durability against cascaded merges ($FBF$).
For cascaded merges, let $\alpha_{\mathrm{eff}}$ denote the final weight fraction from the watermarked model; e.g., two successive merges at $\alpha=0.5$ yield $\alpha_{\mathrm{eff}} \coloneq 0.25$.
We hypothesize that cascaded-merge durability, measured by TPR@1\%, is greater than that of $BB'$ at matched $\alpha_{\mathrm{eff}}$.
If so, $BB'$ provides a lower bound for watermark durability under merging and, because it requires no finetuning, serves as an efficient proxy.

In \cref{fig:lower-bound}, we compare the watermark detectability of $FBF$ with that of $BB'$ at matched $\alpha_{\mathrm{eff}}$.
We find that in the overwhelming majority of settings (93\% of tested configurations), the TPR@1\% of $BB'$ is indeed lower than that of $FBF$, validating our hypothesis.
The lower bound is violated only in the math domain at high $\alpha_{\mathrm{eff}}$ (\ie with a high portion of the math finetuned model).
We suspect this is because math finetuning significantly lowered the entropy of the resulting model on the math domain, making it harder to watermark than even the base (unwatermarked) model.
As a result, a high fraction of the math finetuned model degrades the watermark faster than a similar fraction of the base (unwatermarked) model.
Nonetheless, $BB'$ remains in most cases an effective proxy for evaluating durability against merging.
Full results and the per-$\alpha_{\mathrm{eff}}$ breakdown are in \appref{app:bb-lower-bound}.

\subsection{Ablations on watermark families and model architectures}
\label{sec:experiments:generalization}

\ificmlsubmission
\begin{figure}[t]
\centering
\includegraphics[width=\columnwidth]{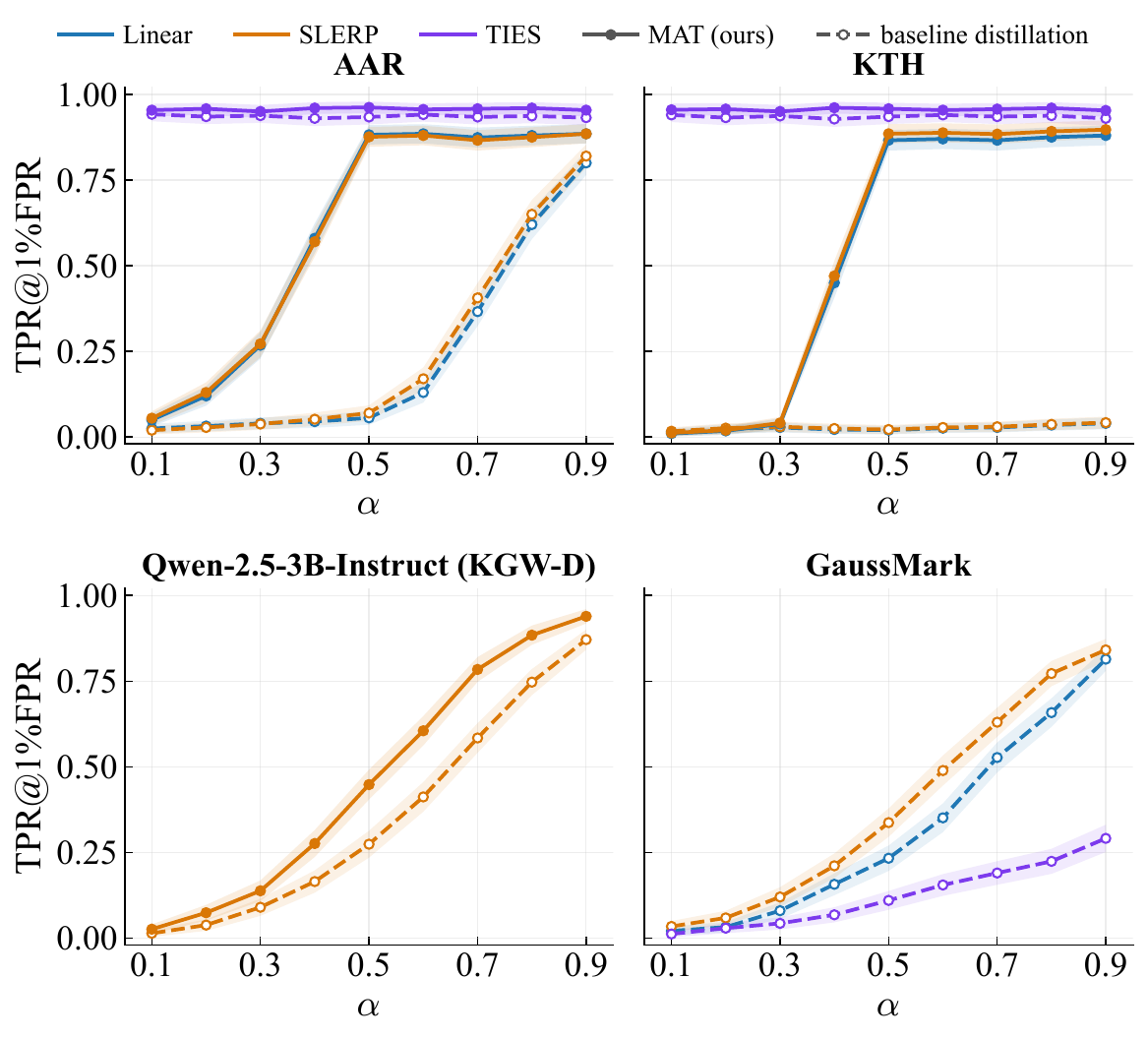}
\caption{\textbf{Durability gain across watermark families and architectures:} Post-merge TPR@1\% FPR under $BB'$ on English (C4). \emph{Top:} \textsc{AAR} (left) and \textsc{KTH} (right) on \textsc{Llama-3.1-8B-Instruct}. \emph{Bottom:} \textsc{KGW-D} on \textsc{Qwen-2.5-3B-Instruct} (left, SLERP only) and \textsc{GaussMark} on \textsc{Llama-3.1-8B-Instruct} (right, baseline only)}
\label{fig:generalization-recovery}
\end{figure}
\else
\begin{wrapfigure}[24]{r}{0.6\textwidth}
    \centering
    \vspace{-1.5em}
    \includegraphics[width=\linewidth]{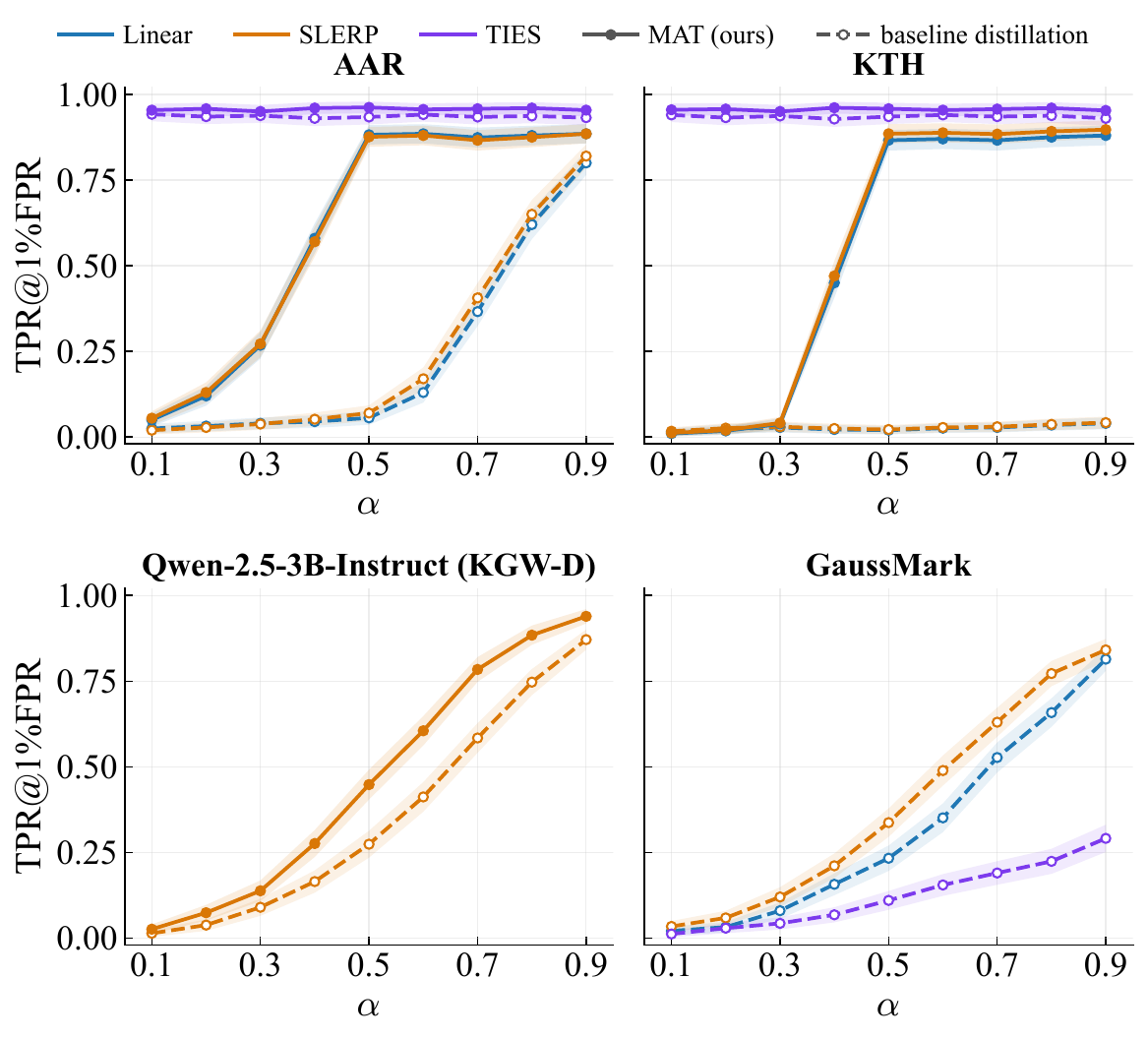}
    \vspace{-2em}
    \caption{\textbf{Durability gain across watermark families and architectures:} Post-merge TPR@1\% FPR under $BB'$ on English (C4). \emph{Top:} \textsc{AAR} (left) and \textsc{KTH} (right) on \textsc{Llama-3.1-8B-Instruct}. \emph{Bottom:} \textsc{KGW-D} on \textsc{Qwen-2.5-3B-Instruct} (left, SLERP only) and \textsc{GaussMark} on \textsc{Llama-3.1-8B-Instruct} (right, baseline only)}
    \label{fig:generalization-recovery}
\end{wrapfigure}
\fi
Next, we test whether our gains are specific to \textsc{KGW-D} on \textsc{Llama-3.1-8B-Instruct} along two distinct axes: watermark family and base architecture.
In particular, we apply the MAT objective to two additional watermark-distillation variants, \textsc{AAR} and \textsc{KTH}, and we also transfer it to a different base model, \textsc{Qwen-2.5-3B-Instruct}.
Additionally, we evaluate the weight-space \textsc{GaussMark}~\citep{gaussmark} on merge robustness in order to verify that merge-induced degradation is not specific to distillation-based OSM watermarking approaches.

\paragraph{MAT generalises across watermark families}
We evaluate durability against $BB'$ across $\alpha$ for \textsc{AAR} and \textsc{KTH}, implemented as in \citep{learnability_wm}, on \textsc{Llama-3.1-8B-Instruct}.
As shown in \cref{fig:generalization-recovery}, baseline distillation collapses under merging at low $\alpha$ for both schemes (TPR@1\% near zero on Linear and SLERP), while our method preserves detectability consistently across the full $\alpha$ range.
Notably, this mirrors the recovery we observe for \kgwd{}, indicating that the merge-adversary objective is not specific to KGW but also transfers to other watermark families. We provide the full results in \appref{app:wm-family-generalization}.

\paragraph{MAT transfers to different base architectures}
We further apply our method to \textsc{Qwen-2.5-3B-Instruct}, reusing the same hyperparameters as in the \textsc{Llama-3.1-8B-Instruct} run.
As we show in \cref{fig:generalization-recovery} (bottom-left), despite this direct transfer, our method maintains a consistent improvement over \kgwd{} across all $\alpha$'s, showing even larger gains for higher $\alpha$'s where the watermarked base model $B$ dominates.
These results indicate that MAT can already provide benefits across base architectures even without requiring model-specific tuning of hyperparameters.

\paragraph{Other watermark schemes also collapse under merging}
Beyond watermark-distillation, we also evaluate \textsc{GaussMark}, which directly embeds the watermark via key-dependent perturbations of the model weights.
The bottom-right panel of \cref{fig:generalization-recovery} shows that \textsc{GaussMark} (like \kgwd{}) collapses under merging, dropping from ${\sim}89\%$ TPR@1\% standalone to ${\sim}34\%$ at $\alpha = 0.5$ using SLERP.
This suggests that the merge-induced collapse is not an artifact of any single watermarking mechanism, motivating future work on adapting MAT to weight-space watermarks.

\section{Limitations and Future Work}
\label{sec:limitations}

While we provide an extensive evaluation of MAT, several directions remain open.
On the merge side, we evaluate MAT across three popular weight-space algorithms (LINEAR, SLERP, TIES), showing that MAT requires only linear interpolation during training.
Although this transfers well to SLERP and TIES, extending MAT to more aggressive trim-and-rescale schemes such as DARE-TIES~\citep{dareties}, Model Breadcrumbs~\citep{davari2024model}, or evolutionary and learned merge operators is an interesting direction for future research.
Similarly, while we are the first to evaluate watermark durability in cascaded scenarios, we focus on depth two to three (\eg $F^{2}F$ and $F^{2}BF$); scaling to longer merge chains that mix more checkpoints is an important direction for real-world model use-cases.
 
On the watermark side, we train MAT variants for \textsc{KGW-D}, \textsc{AAR}, and \textsc{KTH}, and additionally report the merge-collapse behavior of \textsc{GaussMark}.
Adapting MAT to weight-space watermarks like \textsc{GaussMark} would require backpropagating through the detector, which we leave to future work.
More broadly, extending MAT to additional schemes, such as SynthID, semantic watermarks (SemStamp, SemaMark), and unbiased schemes (DiPmark), is a promising avenue for future work.
 
Our main experiments in \cref{sec:experiments} use a single objective formulation with stochastic uniform $\alpha$ and a fixed top-$k$ logit gate.
Although we ablate the chosen parameters in \cref{sec:ablations}, alternative, more complex formulations (\eg curricula over $\alpha$, worst-case-$\alpha$ adversarial training, multi-step inner adversaries, or merge-aware regularizers) may improve the robustness-quality trade-off even further (at the cost of simplicity).
Finally, our experiments focus on \textsc{Llama-3.1-8B-Instruct} and \textsc{Qwen-2.5-3B-Instruct}.
We consider scaling MAT, as in \cref{sec:method}, to larger models conceptually straightforward.
 
\section{Conclusion}
\label{sec:conclusion}
We introduced Merge-Adversarial Training (MAT), a method for improving the durability of open-source model watermarks under model merging.
MAT trains the watermarked model against a simulated merge adversary, encouraging the watermark signal to remain detectable even after the released weights are interpolated with other checkpoints.
Despite using only linear interpolation during training, MAT generalizes across merge algorithms, consistently improving over standard watermark distillation.

We also introduce a broader evaluation pipeline for merge durability and show that unwatermarked-base merges provide an effective worst-case proxy for faster evaluations.
To our knowledge, this is the first work to demonstrate that watermark durability under merging can be explicitly optimized during training without sacrificing model quality.
Together, MAT and our evaluation framework provide a practical step toward operationalizing OSM watermarks that remain detectable after common post-release modifications, and highlight that durability should be treated as a first-class requirement for open-source watermarking alongside detectability, quality, and text-level robustness.

\message{^^JLASTBODYPAGE \thepage^^J}

\clearpage

\bibliographystyle{unsrtnat}
\bibliography{references}
\vfill
\clearpage

\message{^^JLASTREFERENCESPAGE \thepage^^J}

\ifincludeappendixx
	\newpage
	\appendix
	\onecolumn
	\crefalias{section}{appendix}
	\crefalias{subsection}{appendix}
	\begingroup
		\let\OldAppendixSection\section
		\renewcommand{\section}{\FloatBarrier\OldAppendixSection}
		\let\OldAppendixSubsection\subsection
		\renewcommand{\subsection}{\FloatBarrier\OldAppendixSubsection}
		\let\OldAppendixSubsubsection\subsubsection
		\renewcommand{\subsubsection}{\FloatBarrier\OldAppendixSubsubsection}
		\renewcommand{\topfraction}{0.95}
		\renewcommand{\bottomfraction}{0.9}
		\renewcommand{\textfraction}{0.05}
		\renewcommand{\floatpagefraction}{0.7}
		\setcounter{topnumber}{4}
		\setcounter{bottomnumber}{3}
		\setcounter{totalnumber}{6}
		\section{Supplementary Results}
\label{app:supplementary}

\subsection{Watermark Distillation and SFT Hyperparameters}
\label{app:wm-distill}

\subsubsection{Data Mixture}
\label{app:wm-distill:data-mixture}
Watermark distillation uses a multi-domain mix: 30\% math (\textsc{MetaMathQA}~\citep{metamathqa}), 30\% German (20\% \textsc{Alpaca-GPT4-DE}~\citep{alpacagpt4de} + 10\% \textsc{FineWeb-2}~\citep{fineweb2} \texttt{deu\_Latn}), 30\% English (20\% \textsc{Alpaca-GPT4}~\citep{alpacagpt4} + 10\% \textsc{OpenWebText}~\citep{openwebtext}), and 10\% code (\textsc{CodeAlpaca}~\citep{codealpaca}).  Datasets are streamed and interleaved with sampling probabilities equal to the percentages above (fixed seed 42), and each sample is truncated/packed to a sequence length of $512$ tokens.

\subsubsection{Watermark Distillation}
\label{app:wm-distill:training}
Both \textsc{KGW-D} (replication of \citep{learnability_wm}) and \textsc{MAT KGW-D} are trained from \textsc{Llama-3.1-8B-Instruct} \citep{llama3} in \textsc{bfloat16}, using AdamW ($\beta_1{=}0.9,\, \beta_2{=}0.999$, $\varepsilon{=}10^{-8}$), cosine LR schedule with $500$ warm-up steps and peak LR $1{\times}10^{-5}$, a batch size of $64$, and gradient checkpointing enabled.  We use the same KGW parameters in both runs ($\gamma{=}0.25$, $k{=}1$). The bias is $\delta{=}2.3$ for \textsc{KGW-D} and \textsc{MAT KGW-D}, with the latter additionally using the rank-top-$k$ restriction ($k{=}100$).  The merge adversary samples $\alpha\sim\mathcal{U}(0.1,1.0)$ per step and is disabled for the KGW-D baseline. Both \textsc{KGW-D} and \textsc{MAT KGW-D} are trained for $5{,}000$ steps. We run the experiments on NVIDIA RTX Blackwell 6000 Pro GPUs with 96GB VRAM. \textsc{KGW-D} uses $1{\times}4$ GPUs under FSDP (full shard, $\sim 4$ h training time) and \textsc{MAT KGW-D} uses $2{\times}4$ GPUs under DeepSpeed ZeRO-2 ($\sim 6$ h training time). The teacher copy is held on CPU and moved to GPU on demand to fit the merge-adversary forward pass within GPU memory.

\subsubsection{Per-Domain SFT}
\label{app:wm-distill:sft}
Each domain SFT starts from the \textsc{KGW-D}/\textsc{MAT KGW-D} checkpoint-$5{,}000$ and uses AdamW with weight decay $0.01$, cosine schedule and \texttt{bfloat16} with gradient checkpointing. Hyperparameters per domain are reported in \cref{tab:sft-hyperparameters}.

\begin{table}[!htbp]
\centering
\caption{\textbf{Per-domain SFT hyperparameters:} Each finetune starts from the corresponding \textsc{KGW-D}/\textsc{MAT KGW-D} checkpoint-$5{,}000$.}
\label{tab:sft-hyperparameters}
\small
\setlength{\tabcolsep}{4pt}
\begin{tabular}{lcccccc}
\toprule
Domain & Dataset & Epochs & LR & Eff.\ batch & Seq.\ len\\
\midrule
Math    & \textsc{NuminaMath-CoT} \citep{numinamath}   & 1 & $1{\times}10^{-5}$ & $128$ & $4096$  \\
German  & \textsc{Evol-Instruct-DE} \citep{evol-de}          & 2 & $2{\times}10^{-5}$ & $64$  & $2048$  \\
French  & \textsc{Lucie} + \textsc{French-Alpaca} (58k) \citep{lucie, alpaca-fr}                  & 2 & $2{\times}10^{-5}$ & $64$  & $2048$ \\
\bottomrule
\end{tabular}
\end{table}

\noindent The French SFT mixes 29k French-Alpaca instructions sampled from \citep{alpaca-fr} with 29k continued-pretraining pseudo-pairs sampled from the \texttt{RedPajama-fr} subset of the Lucie Training Dataset \citep{lucie} for a total of 58k examples, chosen for sample-count parity with the German \textsc{Evol-Instruct-DE} set so both fine-tunes run for an equal number of optimizer steps. All other settings follow the HuggingFace \texttt{Trainer} defaults. 

\subsection{Benchmark Results and Downstream Capabilities}
\label{app:benchmarks}
\subsubsection{Downstream Capabilities}
\label{app:benchmarks:downstream}
We evaluate downstream capabilities with the EleutherAI \texttt{lm-evaluation-harness}~\citep{eval-harness} using each benchmark's standard few-shot regime: for English/general reasoning we run \textsc{MMLU}~\citep{mmlu} (5-shot), \textsc{HellaSwag}~\citep{hellaswag} (10-shot), \textsc{ARC-Challenge}~\citep{arc} (25-shot). For math we run \textsc{GSM8K}~\citep{gsm8k} (8-shot) and \textsc{Minerva-MATH (MATH)}~\citep{math} (0-shot). Because two of our merge-parent fine-tunes target non-English domains (German and French), we additionally evaluate on \textsc{MMLU-DE}~\citep{mmlu}, \textsc{HellaSwag-DE}~\citep{hellaswag} and the analogous \texttt{fra\_Latn} suite.

\subsubsection{Generation Quality}
\label{app:benchmarks:quality}
We measure generation quality using the reference-model perplexity protocol established by \citet{learnability_wm} and used in \citet{gloaguen2025towards}: from the watermarked model, we generate, unless otherwise mentioned, $n = 500$ continuations conditioned on natural-text prefixes drawn from a domain-matched corpus (\textsc{C4-RealNewsLike} \citep{c4} for English, \textsc{FineWeb-2}~\texttt{deu\_Latn} \citep{fineweb2}/\texttt{fra\_Latn} \citep{fineweb2} for German/French, and \textsc{GSM8K} \citep{gsm8k} prompts for math). We compute their perplexity with \textsc{Llama-2-13B} \citep{llama2} and report the median PPL. Alongside reference perplexity, we report \emph{seq-rep-3}, the fraction of repeated trigrams in the generated continuation, which captures the degenerate repetition failure mode that low perplexity alone can mask.

\subsubsection{Baseline Model Benchmarks}
\label{app:benchmarks:baseline}
\cref{tab:capability-pre-ft} (\textsc{Bases} block) reports the downstream performance of the base \textsc{Llama-3.1-8B-Instruct} model and both watermarked variants (KGW-D and MAT KGW-D) prior to any finetuning or merging. Watermark distillation incurs limited quality loss: reference perplexity rises by ${\sim}0.9$--$1.9$ PPL and MMLU, HellaSwag, and MATH scores remain within $\sim$2\,pp of the unwatermarked baseline, confirming that the watermark embedding itself does not substantially degrade model utility.

\subsubsection{Finetune Benchmarks}
\label{app:benchmarks:finetune}
The \textsc{Math FT}, \textsc{German FT}, and \textsc{French FT} blocks of \cref{tab:capability-pre-ft} compare the per-language finetunes derived from each watermarked variant. All three finetunes achieve comparable on-domain quality regardless of whether the parent model was trained with KGW-D or MAT: German finetunes reach similar perplexity on held-out German text, math finetunes show nearly equivalent MATH accuracy, and French finetunes match on French perplexity and MMLU-FR.

\begin{table}[!htbp]
\centering
\caption{\textbf{Per-task downstream accuracy across the unwatermarked \textsc{Llama-3.1-8B-Instruct} base, the two watermarked bases (\textsc{KGW-D}, \textsc{MAT KGW-D}), and the Math, German, and French finetunes derived from each watermarked base, no merging:} Cells colored relative to the Instruct baseline:
\colorbox{cgreen}{${\geq}95\%$},
\colorbox{corange}{${\geq}90\%$},
\colorbox{cred}{${<}90\%$}.
Per-language perplexity and seq-rep-3 are reported separately in \cref{tab:capability-pre-ft-quality}.}
\footnotesize
\label{tab:capability-pre-ft}
\vspace{4pt}
\small
\setlength{\tabcolsep}{4pt}
\begin{adjustbox}{max width=\textwidth}
\begin{tabular}{ll ccccccc}
\toprule
 & & \multicolumn{7}{c}{Accuracy [\%]} \\
\cmidrule(lr){3-9}
Model & Scenario & ARC & MMLU & HeSw & GSM8K & MATH & MMLU\_DE & MMLU\_FR \\
\midrule
\multirow{3}{*}{\textsc{Bases}}
 & Instruct
   & 61.9 & 68.8 & 76.7 & 85.0 & 44.0 & 57.5 & 59.3 \\
 & KGW-D
   & \cellcolor{corange}58.5 & \cellcolor{cgreen}67.5 & \cellcolor{cgreen}74.7 & \cellcolor{cgreen}82.5 & \cellcolor{cgreen}42.0
   & \cellcolor{cgreen}56.1 & \cellcolor{cgreen}58.5 \\
 & MAT KGW-D
   & \cellcolor{cgreen}59.3 & \cellcolor{cgreen}67.6 & \cellcolor{cgreen}75.8 & \cellcolor{cgreen}81.9 & \cellcolor{cgreen}42.3
   & \cellcolor{cgreen}56.6 & \cellcolor{cgreen}58.9 \\
\midrule
\multirow{2}{*}{\textsc{Math FT}}
 & KGW-D
   & \cellcolor{cred}53.8 & \cellcolor{corange}63.2 & \cellcolor{corange}71.3 & \cellcolor{cgreen}81.9 & \cellcolor{cgreen}47.1
   & \cellcolor{corange}52.3 & \cellcolor{corange}54.1 \\
 & MAT KGW-D
   & \cellcolor{cred}52.7 & \cellcolor{corange}62.1 & \cellcolor{corange}70.5 & \cellcolor{cgreen}82.4 & \cellcolor{cgreen}48.3
   & \cellcolor{cred}51.2 & \cellcolor{corange}55.4 \\
\midrule
\multirow{2}{*}{\textsc{German FT}}
 & KGW-D
   & \cellcolor{corange}56.6 & \cellcolor{corange}63.1 & \cellcolor{corange}70.4 & \cellcolor{corange}76.8 & \cellcolor{cred}36.8
   & \cellcolor{cgreen}58.3 & \cellcolor{corange}53.4 \\
 & MAT KGW-D
   & \cellcolor{corange}56.0 & \cellcolor{corange}62.3 & \cellcolor{corange}70.2 & \cellcolor{cred}75.9 & \cellcolor{cred}37.4
   & \cellcolor{cgreen}57.9 & \cellcolor{corange}54.0 \\
\midrule
\multirow{2}{*}{\textsc{French FT}}
 & KGW-D
   & \cellcolor{cred}52.6 & \cellcolor{cred}60.9 & \cellcolor{corange}71.1 & \cellcolor{cred}74.2 & \cellcolor{cred}37.7
   & \cellcolor{cred}43.1 & \cellcolor{cgreen}58.9 \\
 & MAT KGW-D
   & \cellcolor{cred}51.5 & \cellcolor{cred}54.4 & \cellcolor{corange}70.4 & \cellcolor{cred}75.4 & \cellcolor{cred}36.0
   & \cellcolor{cred}42.1 & \cellcolor{cgreen}58.7 \\
\bottomrule
\end{tabular}
\end{adjustbox}
\end{table}

\begin{table}[!htbp]
\centering
\caption{\textbf{Per-language generation quality (no merging):} Reference perplexity (PPL, computed under \textsc{Llama-2-13B}) and seq-rep-3 of \textsc{Llama-3.1-8B-Instruct} under each watermarking variant. Per-task downstream accuracy is reported separately in \cref{tab:capability-pre-ft}.}
\label{tab:capability-pre-ft-quality}
\vspace{4pt}
\small
\setlength{\tabcolsep}{4pt}
\begin{tabular}{ll ccc ccc}
\toprule
 & & \multicolumn{3}{c}{PPL} & \multicolumn{3}{c}{seq-rep-3} \\
\cmidrule(lr){3-5} \cmidrule(lr){6-8}
Model & Scenario & EN & DE & FR & EN & DE & FR \\
\midrule
\multirow{3}{*}{\textsc{Bases}}
 & Instruct      & 3.81 & 3.98 & 4.55 & 0.058 & 0.093 & 0.077 \\ %
 & KGW-D         & 5.39 & 5.66 & 5.93 & 0.049 & 0.095 & 0.076 \\ %
 & MAT KGW-D     & 5.69 & 5.59 & 5.46 & 0.058 & 0.094 & 0.089 \\ %
\midrule
\multirow{2}{*}{\textsc{Math FT}}
 & KGW-D         & 5.41 & 5.71 & 5.87 & 0.051 & 0.097 & 0.081 \\
 & MAT KGW-D     & 5.70 & 5.73 & 5.40 & 0.059 & 0.096 & 0.090 \\
\midrule
\multirow{2}{*}{\textsc{German FT}}
 & KGW-D         & 5.40 & 3.53 & 5.89 & 0.061 & 0.078 & 0.087 \\
 & MAT KGW-D     & 5.68 & 3.65 & 5.42 & 0.054 & 0.080 & 0.092 \\
\midrule
\multirow{2}{*}{\textsc{French FT}}
 & KGW-D         & 5.43 & 5.69 & 4.09 & 0.048 & 0.091 & 0.072 \\ %
 & MAT KGW-D     & 5.71 & 5.71 & 4.14 & 0.057 & 0.096 & 0.070 \\ %
\bottomrule
\end{tabular}
\end{table}

\begin{table}[!htbp]
\centering
\caption{\textbf{Standalone single-model TPR@1\% FPR detection (no merging):} Bases are evaluated before any finetuning; the finetune blocks are evaluated post-SFT. \textsc{Logit proc.\ KGW} is the decoding-time KGW reference.}
\vspace{4pt}
\label{tab:standalone-detection}
\footnotesize
\begin{tabular}{ll ccc}
\toprule
\textbf{Scenario} & \textbf{Method} & \textbf{English} & \textbf{German} & \textbf{Math} \\
\midrule
\multirow{3}{*}{\textsc{Bases}}
  & Logit proc.\ KGW & 100.0 & 100.0 & 100.0 \\
  & KGW-D            &  98.7 &  98.8 &  95.8 \\
  & MAT KGW-D (ours) &  98.0 &  99.0 &  96.4 \\
\midrule
\multirow{2}{*}{\textsc{German FT}}
  & KGW-D     & 35.4 & 14.0 & 25.0 \\
  & MAT KGW-D & 33.8 & 17.1 & 28.5 \\
\midrule
\multirow{2}{*}{\textsc{Math FT}}
  & KGW-D     & 34.9 & 41.4 & 21.4 \\
  & MAT KGW-D & 36.1 & 39.8 & 24.4 \\
\midrule
\multirow{2}{*}{\textsc{French FT (unseen)}}
  & KGW-D     & 52.0 & 42.2 & 21.0 \\
  & MAT KGW-D & 51.4 & 53.8 & 27.0 \\
\bottomrule
\end{tabular}
\end{table}

\subsection{Supplementary Results for All-Watermarked Merges}
\label{app:all-watermarked}
\Cref{fig:m1-recovery-full}, \cref{fig:m2-recovery}, \cref{fig:m3-recovery}, and \cref{fig:m4-recovery} provide the full TPR@1\% sweeps across the all-watermarked merge family ($FF$, $BF$, $F^{2}BF$, $FBF$), all merge methods (Linear, SLERP, TIES), and all three detection domains (English, German, Math) as recovery curves.
We report 95\% confidence intervals on every TPR@1\% with $n = 500$ prompts per cell. The shaded band around each curve in \cref{fig:m1-recovery-full,fig:m2-recovery,fig:m3-recovery,fig:m4-recovery} and the unwatermarked-parent recovery figures depicts this interval.

\subsubsection{Merge Methods and Hyperparameters}
\label{app:all-watermarked:merge-methods}
We evaluate three merge methods, all parameterised by a single mixing coefficient $\alpha \in [0, 1]$.
\textsc{Linear}~\citep{model_soups} averages the two parents in weight space: $\theta_{\mathrm{merge}} = \alpha\,\theta_A + (1-\alpha)\,\theta_B$.
\textsc{SLERP}~\citep{mergekit} interpolates the two parents on the unit hypersphere with the same coefficient, normalising and renormalising each parameter tensor before/after the spherical interpolation.
\textsc{TIES}~\citep{ties} is computed via \textsc{MergeKit} \citep{mergekit} with parent~$A$ acting as the task-vector base: we form a task vector $\tau_B = \theta_B - \theta_A$, retain the top-$d$ fraction of $|\tau_B|$ per parameter tensor (density $d = 0.5$ throughout), and combine as $\theta_{\mathrm{merge}} = \theta_A + \alpha\,\widetilde{\tau}_B$, where $\widetilde{\tau}_B$ is the magnitude-trimmed and renormalised task vector.
We sweep $\alpha \in \{0.1, 0.2, \ldots, 0.9\}$ for all three methods in the per-merge results in \cref{fig:m1-recovery-full,fig:m2-recovery,fig:m3-recovery,fig:m4-recovery}.

\begin{figure}[!t]
\centering
\includegraphics[width=\textwidth]{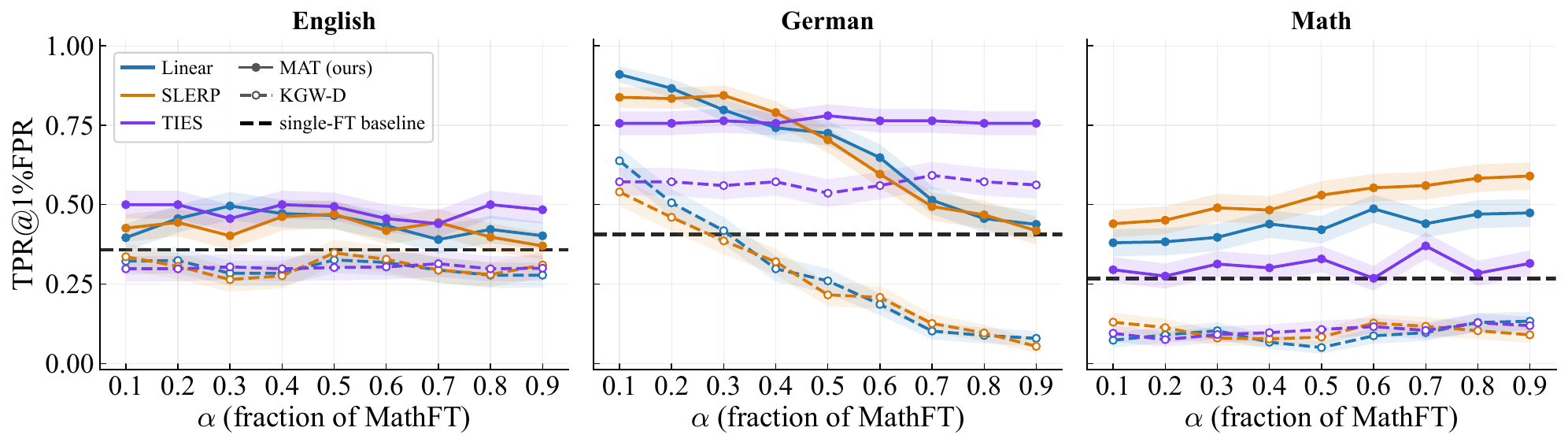}
\caption{\textbf{$FF$ watermark detection:} TPR@1\% as a function of $\alpha$ = fraction of Math FT across Linear, SLERP, and TIES.}
\label{fig:m1-recovery-full}
\end{figure}

\begin{figure}[!t]
\centering
\includegraphics[width=\textwidth]{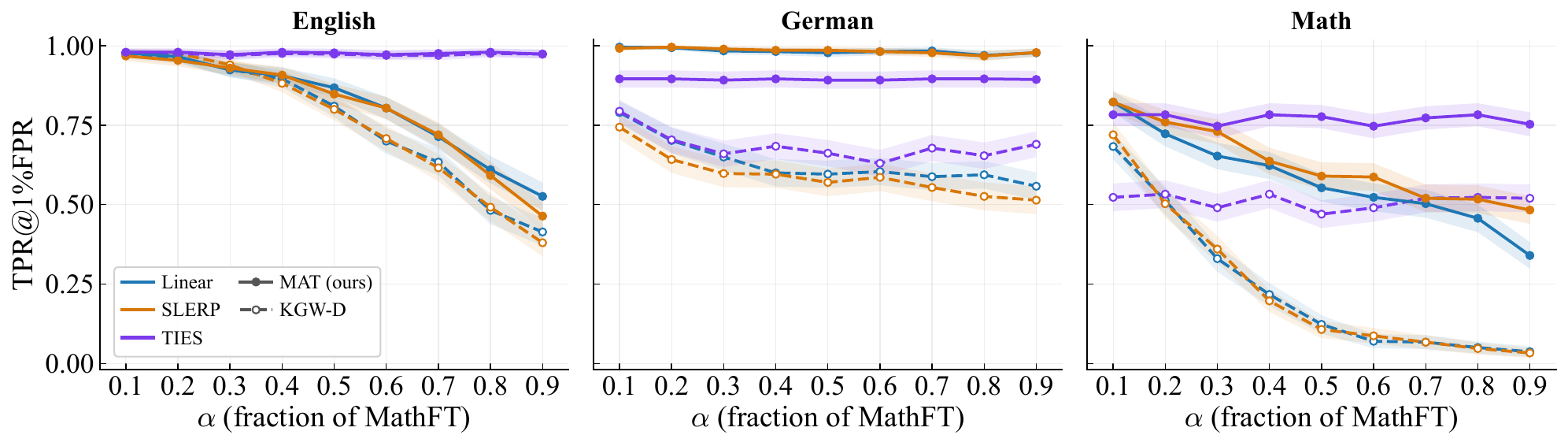}
\caption{\textbf{$BF$ watermark detection:} TPR@1\% as a function of $\alpha$ = fraction of Math FT across Linear, SLERP, and TIES.}
\label{fig:m2-recovery}
\end{figure}

\begin{figure}[!t]
\centering
\includegraphics[width=\textwidth]{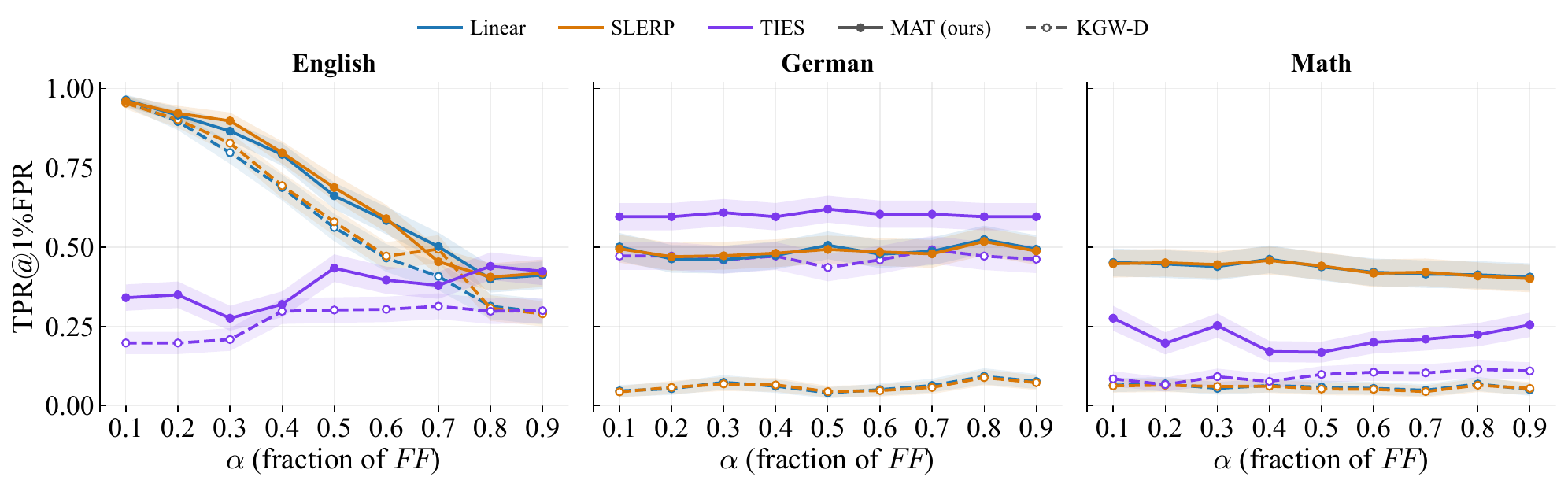}
\caption{\textbf{$F^{2}BF$ watermark detection:} TPR@1\% as a function of $\alpha$ = fraction of $FF$ across Linear, SLERP and TIES.}
\label{fig:m3-recovery}
\end{figure}

\begin{figure}[!t]
\centering
\includegraphics[width=\textwidth]{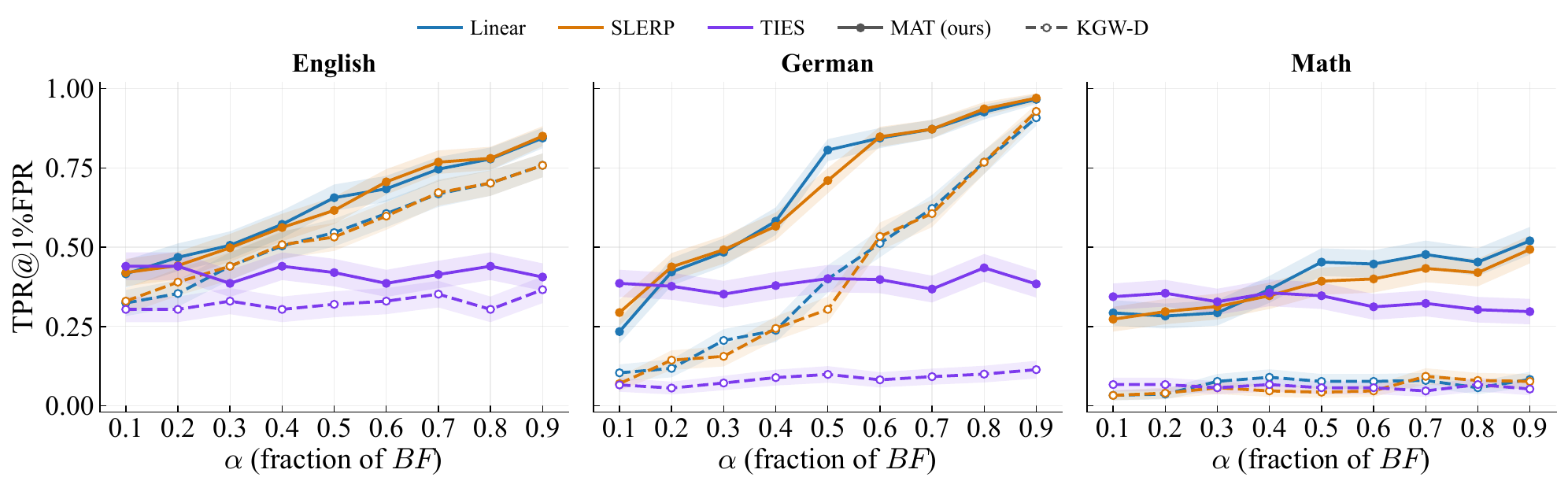}
\caption{\textbf{$FBF$ watermark detection:} TPR@1\% as a function of $\alpha$ = fraction of $BF$ across Linear, SLERP, and TIES.}
\label{fig:m4-recovery}
\end{figure}

\subsubsection{$BB'$ as a Lower Bound for $FBF$}
\label{app:bb-lower-bound}
In \cref{sec:experiments:simplified}, we introduce $BB'$ as an empirical lower bound on $FBF$ at matched effective watermark dose. Here we present the per-$\alpha$ sweep behind that claim (\cref{fig:lower-bound-full}). The matched dose is the B-fraction in the merged weights: $\alpha_{\mathrm{eff}}(FBF) = 0.5 \cdot \alpha_{FBF}$, since $BF$ contains $0.5\,B$. The bound holds where the $FBF$ curve lies above the $BB'$ curve at the same $\alpha_{\mathrm{eff}}$.

\begin{figure}[!t]
\centering
\includegraphics[width=\textwidth]{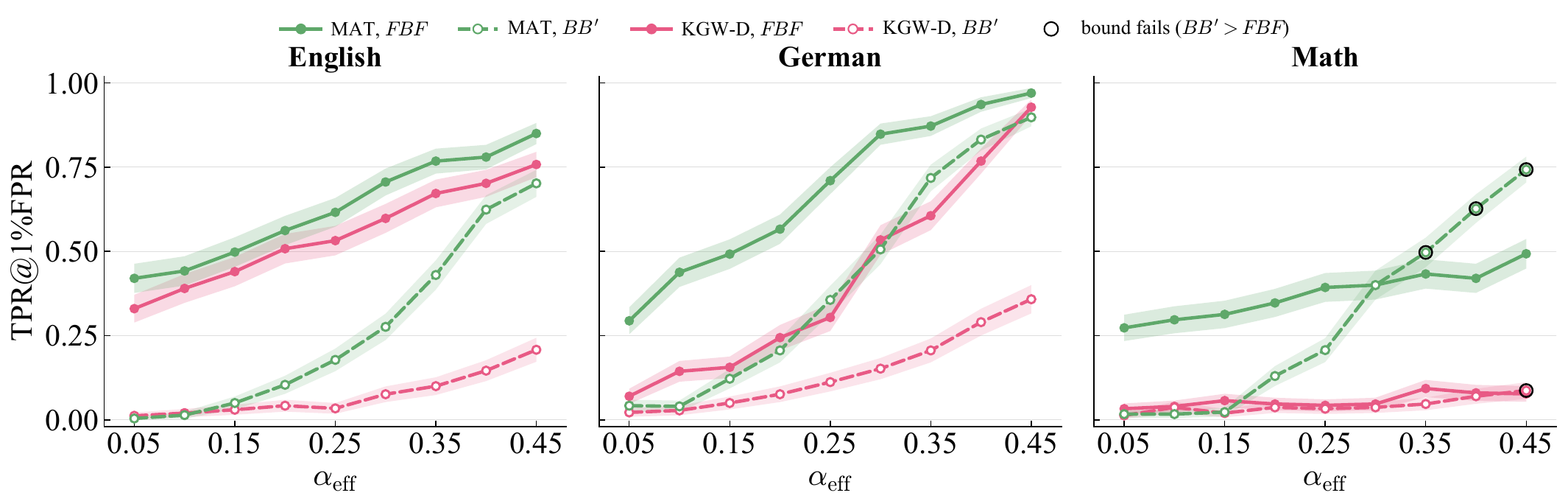}
\caption{\textbf{$FBF$ vs.\ $BB'$ lower-bound test on TPR@1\%, full SLERP sweep $\alpha_{\mathrm{eff}} \in \{0.05, 0.10, \ldots, 0.45\}$:} The bound holds where the solid line sits above the dashed line; configurations where it fails ($BB' > FBF$) are circled.}
\label{fig:lower-bound-full}
\end{figure}

The bound holds across nearly all $(\alpha_{\mathrm{eff}}, \text{domain})$ configurations. The few failures all fall in the Math domain at high $\alpha_{\mathrm{eff}}$: the Math FT parent on its own retains very little watermark signal, so when it dominates the $FBF$ merge it contributes less detectable signal than $BB'$'s simple mix of $B$ with the unwatermarked Instruct. The same conclusion holds under Linear merging (\cref{fig:lower-bound-linear}).

The lower-bound test does not extend to TIES at density $d=0.5$: TIES keeps the dominant per-parameter values from each parent rather than averaging them, so the watermarked $B$ side of $BB'$ is largely preserved and TPR@1\% remains high (${\sim}80$--$90\%$ on English, ${\sim}82$--$91\%$ on German, ${\sim}70$--$88\%$ on Math) regardless of $\alpha_{\mathrm{eff}}$. $BB'$ therefore sits above $FBF$ at every $\alpha_{\mathrm{eff}}$ and the proxy is uninformative; we omit the corresponding figure and instead report the raw values in \cref{tab:lower-bound-ties}.

\begin{table}[!htbp]
\centering
\caption{\textbf{TIES variant of the $FBF$ vs.\ $BB'$ lower-bound test (density $d = 0.5$):} Values are TPR@1\% FPR (in \%) at $\alpha = 0.5$. $BB'$ remains substantially higher than $FBF$ in every cell, so the lower-bound interpretation does not apply on TIES.}
\label{tab:lower-bound-ties}
\small
\setlength{\tabcolsep}{5pt}
\begin{tabular}{l cc cc}
\toprule
& \multicolumn{2}{c}{$FBF$} & \multicolumn{2}{c}{$BB'$} \\
\cmidrule(lr){2-3} \cmidrule(lr){4-5}
\textbf{Domain} & KGW-D & MAT & KGW-D & MAT \\
\midrule
English & 32.0 & 42.0 & 79.7 & 90.0 \\
German  &  9.9 & 40.1 & 81.7 & 91.3 \\
Math    &  5.7 & 34.7 & 70.0 & 87.7 \\
\bottomrule
\end{tabular}
\end{table}

\begin{figure}[!t]
\centering
\includegraphics[width=\textwidth]{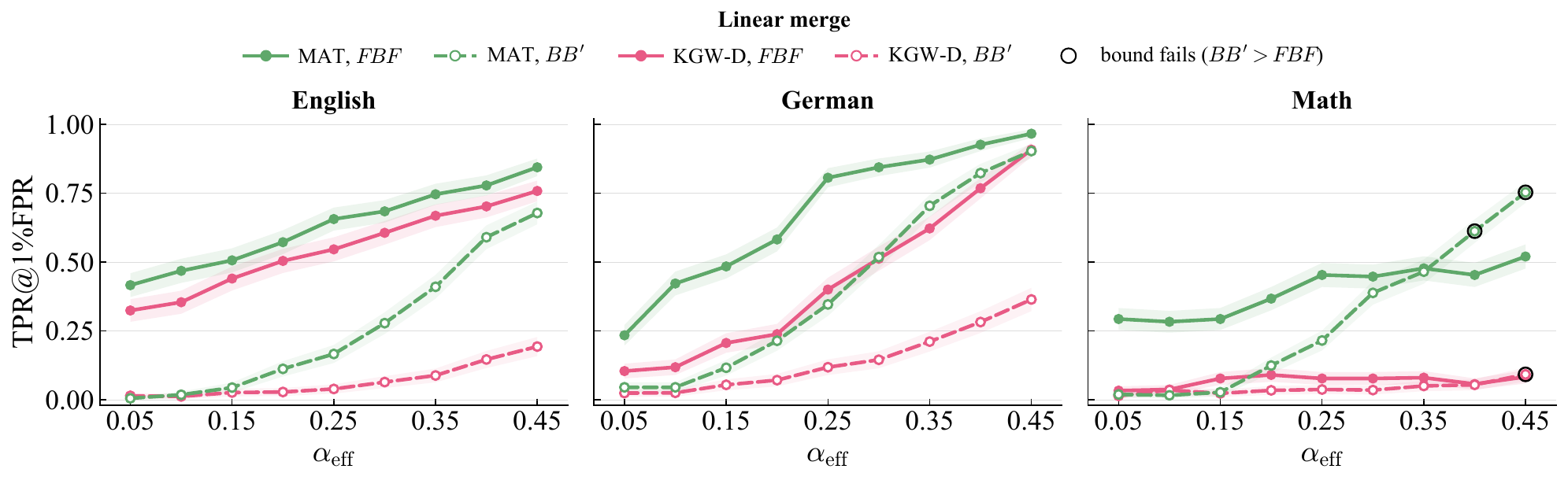}
\caption{\textbf{$FBF$ vs.\ $BB'$ lower-bound test on TPR@1\%, full Linear sweep $\alpha_{\mathrm{eff}} \in \{0.05, 0.10, \ldots, 0.45\}$:} Same conventions as \cref{fig:lower-bound-full}: the bound holds where the solid line sits above the dashed line; configurations where it fails ($BB' > FBF$) are circled.}
\label{fig:lower-bound-linear}
\end{figure}

\subsubsection{$FF$ vs.\ the Single-FT Baseline}
\label{app:all-watermarked:ff-vs-baseline}
\cref{tab:m1-colored-app} compares the merged-model TPR@1\% against the better of the two single-finetune baselines (max of Math FT and German FT, by domain) for both KGW-D and MAT across Linear, SLERP, and TIES at $\alpha \in \{0.3, 0.5, 0.7\}$. MAT sits above the single-FT baseline (\colorbox{cgreen}{green}) in every Linear and SLERP cell, and dominates on the German domain across all three merge methods. On TIES, MAT clears the baseline on English and German across all three $\alpha$, and on Math at $\alpha \in \{0.5, 0.7\}$; the only \colorbox{corange}{orange} cell (TIES Math at $\alpha = 0.3$, within $1$ SE of the baseline) is unsurprising given that the watermark on Math is intrinsically weak post-SFT (${\sim}25$--$29\%$ standalone), leaving little headroom for the merge to add signal. KGW-D, by contrast, falls below or within $1$ standard error in 24 of 27 cells and only exceeds the baseline (\colorbox{cgreen}{green}) on the TIES German rows. This is structural to TIES rather than a merge-recovery effect: TIES keeps the dominant per-parameter values from each parent, so the strong watermark signal in $B$ that survives at standalone is preserved through the merge on the German domain, while Linear and SLERP dilute it. In other words, KGW-D cannot recover the signal beyond the single-finetune baseline under Linear and SLERP, while MAT consistently preserves or recovers it under Linear and SLERP and on the TIES German rows.

\begin{table}[!htbp]
\centering
\caption{\textbf{$FF$ vs.\ single-FT baseline:} $FF$ over Linear, SLERP, TIES at $\alpha \in \{0.3, 0.5, 0.7\}$. Values are TPR@1\% FPR (in \%). Cells coloured relative to the \emph{better} of the two single-finetune baselines per (method, domain) (\ie the max of the \textsc{Math FT} and \textsc{German FT} rows in \cref{tab:standalone-detection}): \colorbox{cgreen}{above by $>1$ SE}, \colorbox{corange}{within $\pm 1$ SE}, \colorbox{cred}{below by $>1$ SE}, where SE is the per-column binomial standard error of TPR@1\% on $n=500$ prompts.}
\label{tab:m1-colored-app}
\begin{adjustbox}{max width=\textwidth}
\begin{tabular}{ll cc cc cc}
\toprule
& & \multicolumn{2}{c}{\textbf{English}} & \multicolumn{2}{c}{\textbf{German}} & \multicolumn{2}{c}{\textbf{Math}} \\
\cmidrule(lr){3-4} \cmidrule(lr){5-6} \cmidrule(lr){7-8}
\textbf{Method} & $\alpha$ & KGW-D & MAT & KGW-D & MAT & KGW-D & MAT \\
\midrule
\multirow{3}{*}{Linear}
& 0.3 & \cellcolor{cred}28.4    & \cellcolor{cgreen}49.6    & \cellcolor{corange}41.8 & \cellcolor{cgreen}79.8    & \cellcolor{cred}10.3    & \cellcolor{cgreen}39.7    \\
& 0.5 & \cellcolor{cred}32.6    & \cellcolor{cgreen}46.6    & \cellcolor{cred}26.0    & \cellcolor{cgreen}72.5    & \cellcolor{cred}5.0     & \cellcolor{cgreen}42.1    \\
& 0.7 & \cellcolor{cred}29.4    & \cellcolor{cgreen}39.0    & \cellcolor{cred}10.2    & \cellcolor{cgreen}51.4    & \cellcolor{cred}9.7     & \cellcolor{cgreen}44.0    \\
\midrule
\multirow{3}{*}{SLERP}
& 0.3 & \cellcolor{cred}26.4    & \cellcolor{cgreen}40.2    & \cellcolor{corange}38.6 & \cellcolor{cgreen}84.4    & \cellcolor{cred}8.0     & \cellcolor{cgreen}49.0    \\
& 0.5 & \cellcolor{corange}34.8 & \cellcolor{cgreen}47.0    & \cellcolor{cred}21.6    & \cellcolor{cgreen}70.4    & \cellcolor{cred}8.3     & \cellcolor{cgreen}53.0    \\
& 0.7 & \cellcolor{cred}29.4    & \cellcolor{cgreen}44.4    & \cellcolor{cred}12.6    & \cellcolor{cgreen}49.4    & \cellcolor{cred}11.7    & \cellcolor{cgreen}56.0    \\
\midrule
\multirow{3}{*}{TIES}
& 0.3 & \cellcolor{cred}30.4    & \cellcolor{cgreen}45.6    & \cellcolor{cgreen}56.0  & \cellcolor{cgreen}76.4    & \cellcolor{cred}9.1     & \cellcolor{corange}31.3   \\
& 0.5 & \cellcolor{cred}30.2    & \cellcolor{cgreen}49.4    & \cellcolor{cgreen}53.6  & \cellcolor{cgreen}78.0    & \cellcolor{cred}10.7    & \cellcolor{cgreen}32.9    \\
& 0.7 & \cellcolor{cred}31.4    & \cellcolor{cgreen}44.0    & \cellcolor{cgreen}59.2  & \cellcolor{cgreen}76.4    & \cellcolor{cred}10.4    & \cellcolor{cgreen}37.0    \\
\bottomrule
\end{tabular}
\end{adjustbox}
\end{table}

\subsubsection{Unwatermarked-Parent Merges ($BF'$ and $BB'$)}
\label{app:unwm-parent-merges}
\Cref{fig:bf-fusechat-recovery,fig:bf-openmath-recovery,fig:bf-tulu-recovery} report the per-domain $BF'$ recovery sweeps for the three community finetunes (FuseChat \citep{fusechat}, OpenMath \citep{openmath}, Tulu \citep{tulu}) and \cref{fig:bb-recovery} the $BB'$ recovery against the unwatermarked \textsc{Llama-3.1-8B-Instruct} base. Under Linear and SLERP, MAT keeps the watermark detectable down to smaller fractions of the watermarked base $B$ than KGW-D in every domain. TIES behaves differently: on $BF'$ it remains in the moderately-detectable regime for both methods (KGW-D ${\sim}40$--$76\%$, MAT ${\sim}70$--$86\%$ across domains), with MAT consistently $\sim$8--12\,pp above KGW-D; on $BB'$ TIES stays high but no longer fully saturated, with KGW-D at ${\sim}70$--$82\%$ and MAT at ${\sim}88$--$91\%$.

\begin{figure}[!t]
\centering
\includegraphics[width=\textwidth]{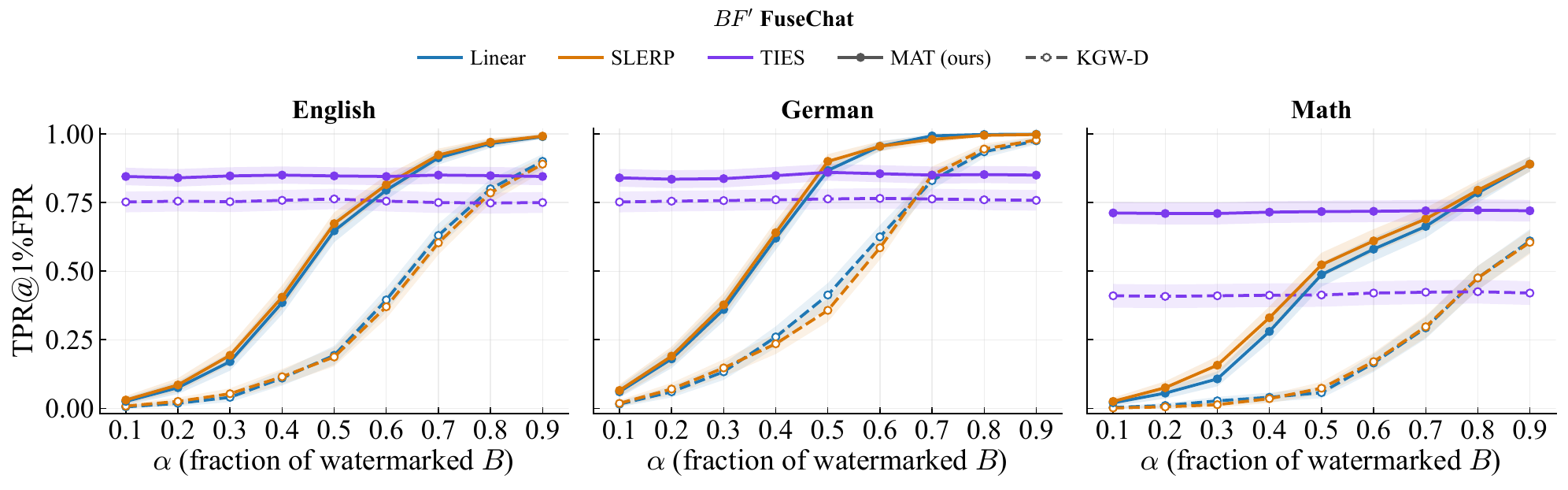}
\caption{\textbf{$BF'$ FuseChat \citep{fusechat} recovery:} TPR@1\% as a function of $\alpha$ = fraction of the watermarked $B$ in $B \otimes \text{FuseChat}$, where FuseChat is a community finetune of \textsc{Llama-3.1-8B-Instruct} \citep{llama3}.}
\label{fig:bf-fusechat-recovery}
\end{figure}

\begin{figure}[!t]
\centering
\includegraphics[width=\textwidth]{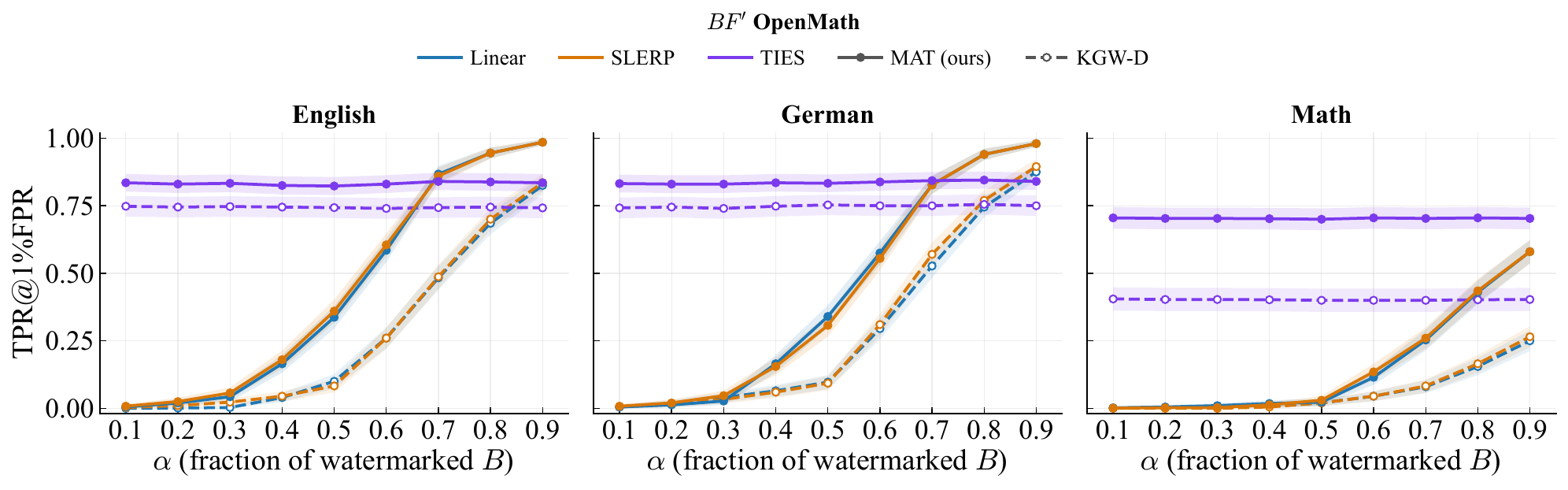}
\caption{\textbf{$BF'$ OpenMath \citep{openmath} recovery:} TPR@1\% as a function of $\alpha$ = fraction of the watermarked $B$ in $B \otimes \text{OpenMath}$, where OpenMath is a community finetune of the \textsc{Llama-3.1-8B} base model \citep{llama3}. Same conventions as \cref{fig:bf-fusechat-recovery}.}
\label{fig:bf-openmath-recovery}
\end{figure}

\begin{figure}[!t]
\centering
\includegraphics[width=\textwidth]{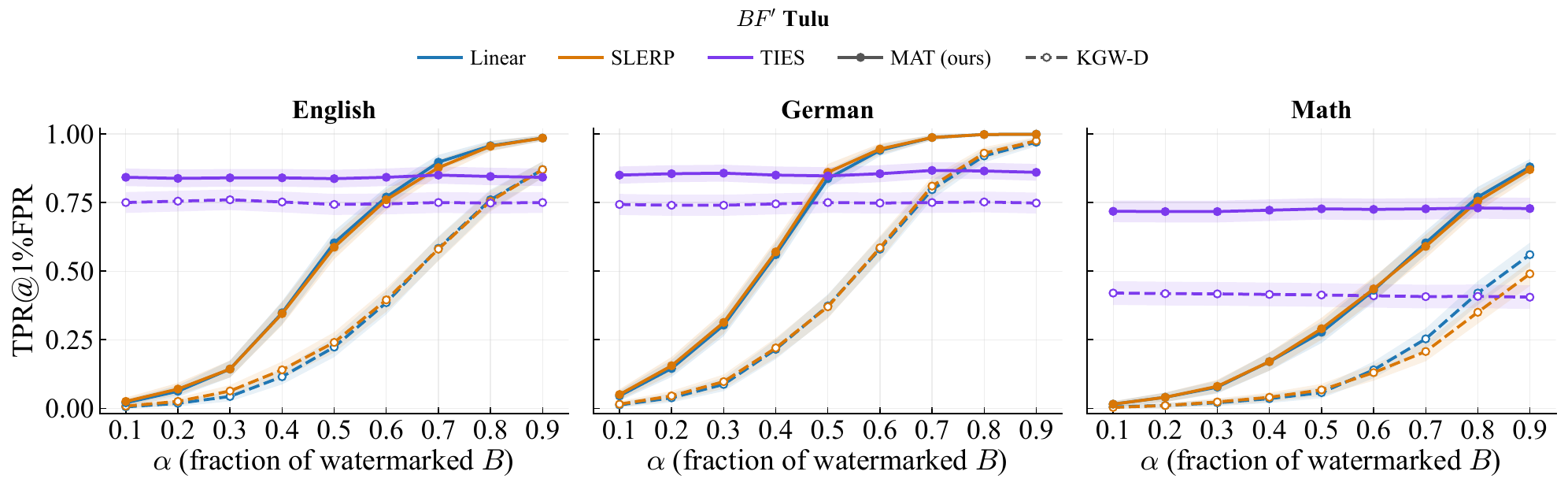}
\caption{\textbf{$BF'$ Tulu \citep{tulu} recovery:} TPR@1\% as a function of $\alpha$ = fraction of the watermarked $B$ in $B \otimes \text{Tulu}$, where Tulu is a community finetune of the \textsc{Llama-3.1-8B} base model \citep{llama3}. Same conventions as \cref{fig:bf-fusechat-recovery}.}
\label{fig:bf-tulu-recovery}
\end{figure}

\begin{figure}[t]
\centering
\includegraphics[width=\textwidth]{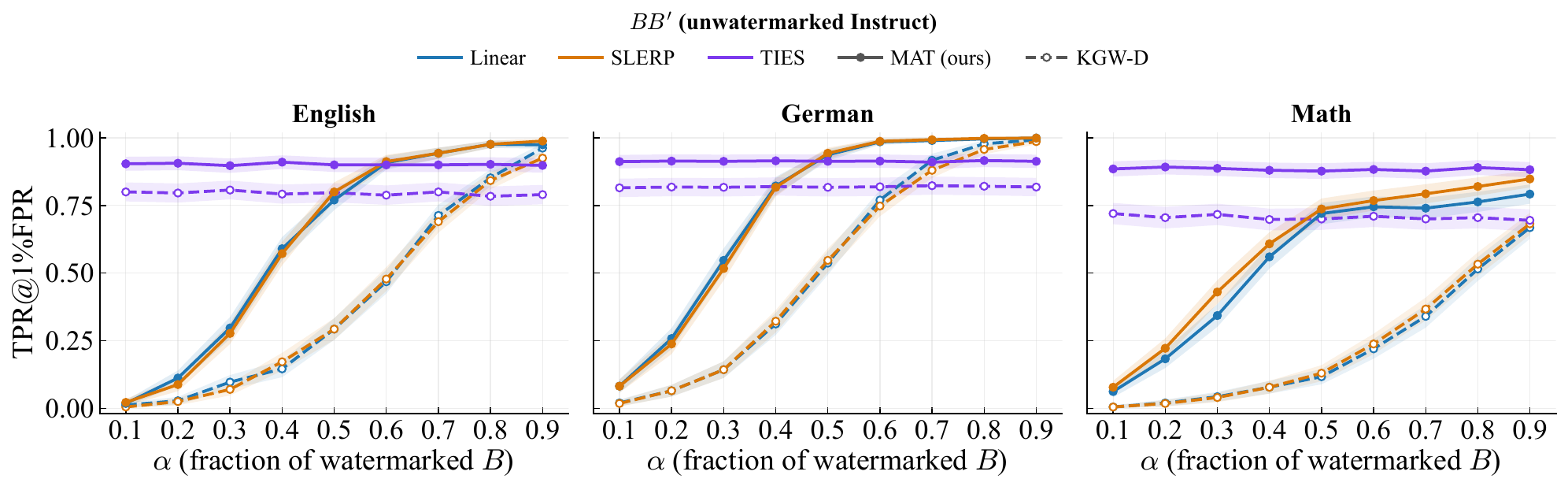}
\caption{\textbf{$BB'$ recovery:} TPR@1\% as a function of $\alpha$ = fraction of the watermarked $B$ in $B \otimes B'_{\text{unwm}}$, where $B'_{\text{unwm}}$ is the unwatermarked \textsc{Llama-3.1-8B-Instruct} base \citep{llama3}.}
\label{fig:bb-recovery}
\end{figure}

\subsubsection{Post-Merge Benchmark Results for $FF$ and $BF$}
\label{app:ff-bf-post-merge-benchmarks}
\cref{tab:capability-pre-ft} reports lm-eval-harness accuracies for the unwatermarked base, the two watermarked bases (\textsc{KGW-D}, \textsc{MAT KGW-D}), and the per-domain finetunes \emph{prior to merging}. \cref{tab:ff-bf-post-merge-benchmarks} reports the same suite of benchmarks on the \emph{merged} checkpoints in the two main all-watermarked merge configurations from \cref{sec:experiments:realistic}: $FF$ (Math FT $\otimes$ German FT) and $BF$ (the watermarked base $B$ paired with each per-domain finetune). The objective is to verify that our merge-adversary objective does not degrade downstream capability of the merged checkpoint relative to the \textsc{KGW-D} baseline at matched merge configuration.

\begin{table}[!htbp]
\centering
\caption{\textbf{Post-merge benchmark results for $FF$ and $BF$:} Per-task downstream accuracy and per-language reference perplexity of the merged models for the $FF$ and $BF$ scenarios at SLERP $\alpha \in \{0.5, 0.7\}$, for both \textsc{KGW-D} and \textsc{MAT KGW-D}. Accuracy cells are coloured relative to the unwatermarked Instruct baseline (\cref{tab:capability-pre-ft}, \textsc{Bases} block):
\colorbox{cgreen}{${\geq}95\%$},
\colorbox{corange}{${\geq}90\%$},
\colorbox{cred}{${<}90\%$}.
PPL is measured under the \textsc{Llama-2-13B} reference model on the same per-language corpora used in \cref{tab:capability-pre-ft-quality}}
\footnotesize
\label{tab:ff-bf-post-merge-benchmarks}
\vspace{4pt}
\small
\setlength{\tabcolsep}{4pt}
\begin{adjustbox}{max width=\textwidth}
\begin{tabular}{lll ccccccc ccc}
\toprule
 & & & \multicolumn{7}{c}{Accuracy [\%]} & \multicolumn{3}{c}{PPL} \\
\cmidrule(lr){4-10} \cmidrule(lr){11-13}
Scenario & $\alpha$ & Variant & ARC & MMLU & HeSw & GSM8K & MATH & MMLU\_DE & MMLU\_FR & EN & DE & FR \\
\midrule
\multirow{4}{*}{$FF$ (MathFT $\otimes$ GermanFT)}
 & \multirow{2}{*}{0.5}
   & \textsc{KGW-D}
   & \cellcolor{corange}56.7 & \cellcolor{cgreen}65.5 & \cellcolor{corange}71.4 & \cellcolor{corange}76.7 & \cellcolor{cgreen}42.0
   & \cellcolor{cgreen}54.7 & \cellcolor{corange}53.8 & 5.42 & 4.65 & 5.86 \\
 & & \textsc{MAT KGW-D}
   & \cellcolor{corange}56.7 & \cellcolor{corange}64.8 & \cellcolor{corange}71.1 & \cellcolor{corange}76.5 & \cellcolor{cgreen}43.0
   & \cellcolor{cgreen}54.7 & \cellcolor{corange}54.7 & 5.68 & 4.55 & 5.42 \\
\cmidrule(l){2-13}
 & \multirow{2}{*}{0.7}
   & \textsc{KGW-D}
   & \cellcolor{corange}55.8 & \cellcolor{corange}65.2 & \cellcolor{corange}70.9 & \cellcolor{corange}77.7 & \cellcolor{cgreen}44.0
   & \cellcolor{cgreen}54.7 & \cellcolor{corange}53.5 & 5.41 & 5.06 & 5.88 \\
 & & \textsc{MAT KGW-D}
   & \cellcolor{corange}55.8 & \cellcolor{corange}64.5 & \cellcolor{corange}70.5 & \cellcolor{corange}77.0 & \cellcolor{cgreen}45.0
   & \cellcolor{cgreen}54.7 & \cellcolor{corange}54.0 & 5.69 & 5.11 & 5.41 \\
\midrule
\multirow{4}{*}{$BF$ (B $\otimes$ MathFT)}
 & \multirow{2}{*}{0.5}
   & \textsc{KGW-D}
   & \cellcolor{corange}56.2 & \cellcolor{corange}65.3 & \cellcolor{corange}72.7 & \cellcolor{cgreen}82.1 & \cellcolor{cgreen}44.5
   & \cellcolor{corange}54.3 & \cellcolor{cgreen}56.4 & 5.40 & 5.68 & 5.90 \\
 & & \textsc{MAT KGW-D}
   & \cellcolor{corange}56.0 & \cellcolor{corange}64.8 & \cellcolor{cgreen}73.0 & \cellcolor{cgreen}82.4 & \cellcolor{cgreen}45.5
   & \cellcolor{corange}53.9 & \cellcolor{cgreen}57.0 & 5.69 & 5.65 & 5.45 \\
\cmidrule(l){2-13}
 & \multirow{2}{*}{0.7}
   & \textsc{KGW-D}
   & \cellcolor{cred}55.7 & \cellcolor{corange}64.5 & \cellcolor{corange}72.2 & \cellcolor{cgreen}82.1 & \cellcolor{cgreen}45.9
   & \cellcolor{corange}53.4 & \cellcolor{corange}55.4 & 5.40 & 5.70 & 5.89 \\
 & & \textsc{MAT KGW-D}
   & \cellcolor{corange}55.9 & \cellcolor{corange}64.0 & \cellcolor{corange}72.1 & \cellcolor{cgreen}82.3 & \cellcolor{cgreen}46.9
   & \cellcolor{corange}53.3 & \cellcolor{cgreen}56.5 & 5.70 & 5.69 & 5.42 \\
\bottomrule
\end{tabular}
\end{adjustbox}
\end{table}

\subsection{Generalization Across Watermark Families and Architectures}
\label{app:wm-family-generalization}

\subsubsection{Watermark Families}

\cref{tab:wm-family-ablation} summarises standalone (un-merged) detection and quality for the AAR and KTH watermark variants on English. \cref{fig:aar-recovery} and \cref{fig:kth-recovery} show the post-merge TPR@1\% recovery on English under the $BB'$ merge across Linear, SLERP, and TIES for AAR and KTH, respectively: in both schemes, the baseline distillation (\textsc{AAR-D}, \textsc{KTH-D}), distilled as in \citep{learnability_wm}, collapses post-merge while MAT preserves detectability across $\alpha$. For \textsc{GaussMark} \citep{gaussmark} we report the baseline detection collapse under merging and discuss it separately in \appref{app:gaussmark}. The \textsc{Qwen-2.5-3B-Instruct} \citep{qwen} ablation under $BB'$ is reported separately in \cref{tab:wm-family-qwen}.

Both \textsc{AAR-D} and \textsc{KTH-D} reproduce the merge-collapse pattern we observed for \textsc{KGW-D}: at $\alpha = 0.5$ under SLERP / Linear $BB'$, post-merge TPR@1\% drops from ${\sim}87$--$90\%$ standalone to ${\sim}7\%$ (\textsc{AAR-D}) and ${\sim}2\%$ (\textsc{KTH-D}). For \textsc{MAT AAR} and \textsc{MAT KTH} we apply only the merge-adversary training on top of the standard distillation; the top-$k$ logit restriction we use for \textsc{MAT KGW-D} is not applied here. Even without that additional component, the merge-adversary objective alone recovers detection to ${\sim}88\%$ (\textsc{MAT AAR}) and ${\sim}87\%$ (\textsc{MAT KTH}) at $\alpha = 0.5$, a $+80$ to $+85$\,pp gain, while leaving standalone detection and generation quality essentially unchanged (\cref{tab:wm-family-ablation}). The merge-induced collapse is therefore not specific to \textsc{KGW} and the merge-adversary objective transfers to other watermark families.

\begin{table}[!htbp]
\centering
\caption{\textbf{Standalone watermark-family ablation on \textsc{Llama-3.1-8B-Instruct}, English (C4):} Pre-merge detection and quality for the AAR and KTH watermark variants. Post-merge results under $BB'$ across Linear, SLERP, and TIES are in \cref{fig:aar-recovery} (AAR) and \cref{fig:kth-recovery} (KTH). GaussMark is reported separately in \appref{app:gaussmark}.}
\vspace{4pt}
\label{tab:wm-family-ablation}
\footnotesize
\setlength{\tabcolsep}{4pt}
\begin{tabular}{l l c cc}
\toprule
& & \textbf{Detection} & \multicolumn{2}{c}{\textbf{Quality}} \\
\cmidrule(lr){3-3} \cmidrule(lr){4-5}
\textbf{Scheme} & \textbf{Variant} & TPR@1\% & PPL & rep-3 \\
\midrule
\multirow{2}{*}{\textsc{AAR}}
 & \textsc{AAR-D}                  & 87.7 & 5.27 & 0.073 \\
 & \textsc{MAT AAR} (ours)         & 84.7 & 5.68 & 0.093 \\
\midrule
\multirow{2}{*}{\textsc{KTH}}
 & \textsc{KTH-D}                  & 89.7 & 6.8  & 0.090 \\
 & \textsc{MAT KTH} (ours)         & 88.5 & 6.2  & 0.097 \\
\bottomrule
\end{tabular}
\end{table}

\begin{figure}[!htbp]
\centering
\includegraphics[width=0.6\textwidth]{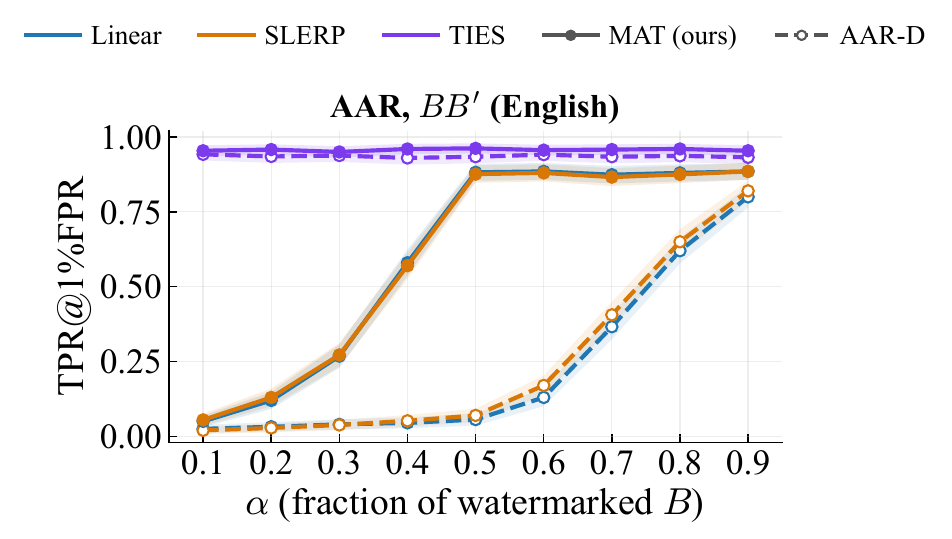}
\caption{\textbf{AAR $BB'$ scenario on English domain:} TPR@1\% as a function of $\alpha$ = fraction of the watermarked AAR base $B$ in $BB'$, across Linear, SLERP, and TIES. \textsc{AAR-D} (dashed, open markers) collapses on Linear/SLERP at $\alpha \leq 0.5$; \textsc{MAT AAR} preserves detectability down to $\alpha = 0.3$. TIES is saturated for both variants.}
\label{fig:aar-recovery}
\end{figure}

\begin{figure}[!htbp]
\centering
\includegraphics[width=0.6\textwidth]{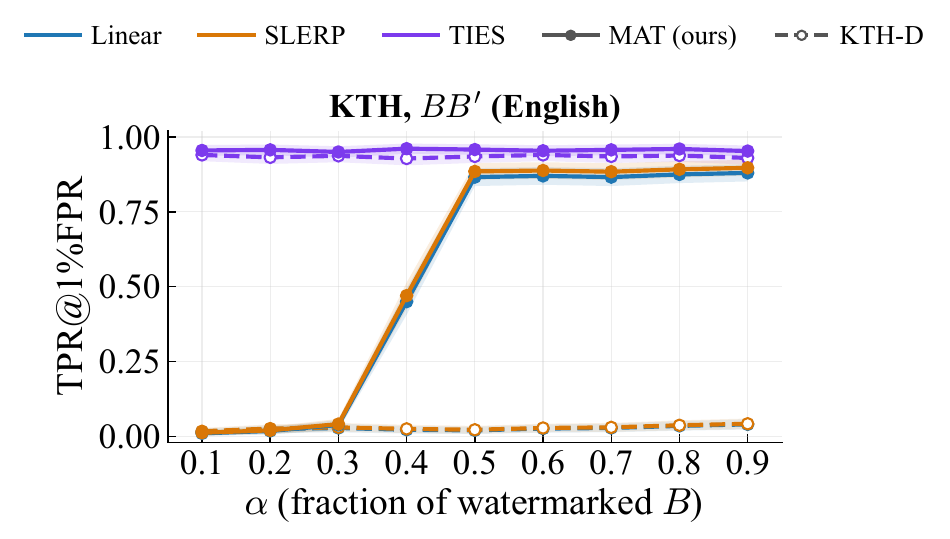}
\caption{\textbf{KTH $BB'$ scenario on English domain:} TPR@1\% as a function of $\alpha$ = fraction of the watermarked KTH base $B$ in $BB'$, across Linear, SLERP, and TIES. \textsc{KTH-D} (dashed, open markers) collapses on Linear/SLERP at $\alpha \leq 0.4$; \textsc{MAT KTH} (solid, filled markers) preserves detectability from $\alpha = 0.4$ upward. TIES is saturated for both variants.}
\label{fig:kth-recovery}
\end{figure}

\subsubsection{Qwen-2.5-3B-Instruct Model}
\cref{tab:wm-family-qwen} reports the same merge-recovery effect on a different base architecture: \textsc{Qwen-2.5-3B-Instruct} \citep{qwen} ($\delta=2.3$, $\gamma=0.25$; logit-processor reference 100\% TPR@1\%, PPL\,=\,11.67, rep-3\,=\,0.041), evaluated on English (C4) under the $BB'$ merge with the unwatermarked \textsc{Qwen-2.5-3B-Instruct}.

We did not retune any of the MAT hyperparameters when porting them from \textsc{Llama-3.1-8B-Instruct}: the same $\delta$, $\gamma$, top-$k$ restriction, and merge-adversary $\alpha$ schedule were applied directly. Even with this transfer, \textsc{MAT KGW-D} improves post-merge English TPR@1\% under $BB'$ by $+17$ to $+20$\,pp at $\alpha \in \{0.5, 0.7\}$ over the \textsc{KGW-D} baseline (\cref{tab:wm-family-qwen}). The gain shrinks at $\alpha = 0.3$ where the unwatermarked Qwen parent dominates the merge, but the qualitative recovery pattern matches the \textsc{Llama} base, suggesting the merge-adversary objective does not require base-specific tuning to provide a benefit.

\begin{table}[!htbp]
\centering
\caption{\textbf{Qwen-2.5-3B-Instruct $BB'$ merge ablation:} TPR@1\% under SLERP $BB'$ across $\alpha \in \{0.3, 0.5, 0.7\}$ on English (C4). \textsc{KGW-D} = standard KGW logit distillation applied to Qwen-2.5-3B-Instruct; \textsc{MAT KGW-D} = the same with merge-adversary training and top-$k$. $\Delta$ = MAT KGW-D $-$ KGW-D (in pp); cells coloured \colorbox{cgreen}{${\geq}{+}10$\,pp}, \colorbox{corange}{${\geq}{+}3$\,pp}, uncolored ${<}{+}3$\,pp.}
\vspace{4pt}
\label{tab:wm-family-qwen}
\footnotesize
\setlength{\tabcolsep}{4pt}
\begin{tabular}{l c rrr}
\toprule
& & \multicolumn{2}{c}{\textbf{TPR@1\%}} & \\
\cmidrule(lr){3-4}
\textbf{Merge} & $\alpha$ & \textsc{KGW-D} & \textsc{MAT KGW-D} (ours) & $\Delta$ \\
\midrule
\multirow{3}{*}{$BB'$ (SLERP)}
  & 0.3 &  9.0 & 13.8 & \cellcolor{corange}\textbf{+4.8}  \\
  & 0.5 & 27.4 & 44.8 & \cellcolor{cgreen}\textbf{+17.4} \\
  & 0.7 & 58.4 & 78.4 & \cellcolor{cgreen}\textbf{+20.0} \\
\bottomrule
\end{tabular}
\end{table}

\subsubsection{GaussMark}
\label{app:gaussmark}
\textsc{GaussMark} \citep{gaussmark} is a weight-space watermark: instead of biasing logits or modifying sampling, it perturbs a fixed subset of the model parameters with low-amplitude Gaussian noise drawn from a key-dependent distribution, and detection runs a likelihood-ratio test on those parameters.

We apply \textsc{GaussMark} to \textsc{Llama-3.1-8B-Instruct} as described in \citet{gaussmark}, and then run the same $BB'$ merge sweep used for \textsc{KGW-D}. Standalone (no merge), the watermarked model reaches ${\sim}89\%$ TPR@1\% on English (\cref{tab:gaussmark-standalone}). Under $BB'$ at $\alpha = 0.5$ under SLERP, it drops to ${\sim}34\%$ TPR@1\% (\cref{fig:gaussmark-recovery}), the same collapse pattern we observe for \textsc{KGW-D}, \textsc{AAR-D}, and \textsc{KTH-D}.

The GaussMark panel in \cref{fig:generalization-recovery} reports the baseline collapse and is included to underline that the collapse is not an artefact of any single watermarking mechanism. 

\begin{table}[!htbp]
\centering
\caption{\textbf{Standalone GaussMark ablation on \textsc{Llama-3.1-8B-Instruct}, English (C4):}}
\vspace{4pt}
\label{tab:gaussmark-standalone}
\footnotesize
\setlength{\tabcolsep}{4pt}
\begin{tabular}{l c cc}
\toprule
& \textbf{Detection} & \multicolumn{2}{c}{\textbf{Quality}} \\
\cmidrule(lr){2-2} \cmidrule(lr){3-4}
\textbf{Variant} & TPR@1\% & PPL & rep-3 \\
\midrule
\textsc{GaussMark} & 89.0 & 4.06 & 0.061 \\
\bottomrule
\end{tabular}
\end{table}

\begin{figure}[!htbp]
\centering
\includegraphics[width=0.6\textwidth]{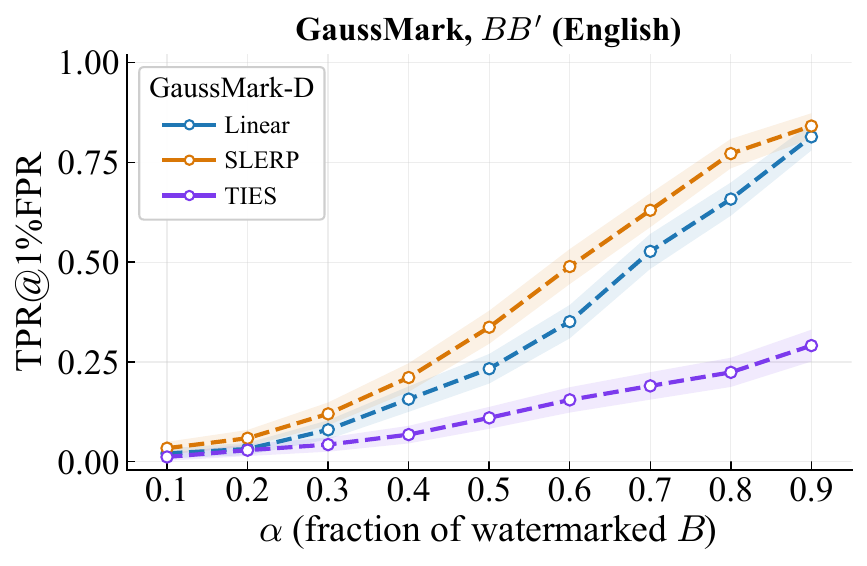}
\caption{\textbf{GaussMark $BB'$ merge scenario on English domain:} TPR@1\% under the $BB'$ merge as a function of $\alpha$ = fraction of the watermarked GaussMark base $B$, across Linear, SLERP, and TIES.}
\label{fig:gaussmark-recovery}
\end{figure}

\section{Ablations}
\label{sec:ablations}

In this section, we ablate the components of our method presented in~\cref{sec:method:training}. 

\paragraph{Experimental Setup}
All ablations are run on \textsc{Llama-3.2-1B-Instruct} for $7\,500$ steps (LR\,=\,$1\mathrm{e}{-}5$, $500$-step warm-up, cosine decay). 
The Hyperparameters are otherwise identical to the full \textsc{Llama-3.1-8B-Instruct} training described in \cref{app:wm-distill}. 
We vary one hyperparameter at a time from the baseline settings $(k\!=\!200,\,\alpha_{\min}\!=\!0.1,\,\delta\!=\!2.0)$ and then validate the joint best in a second pass. 
The hyperparameters used in \cref{sec:method} correspond to the bolded row in each table: $\delta\!=\!2.3$, $k\!=\!100$, $\alpha_{\min}\!=\!0.1$. 
Each row reports standalone (un-merged) TPR@1\% on C4 with PPL and seq-rep-3, alongside post-merge TPR@1\% under Linear merging at $\alpha\!=\!0.5$ as the headline robustness number.

\subsection{Top-$k$ Gating ($k$)}
\label{sec:ablations:top-k}
$k$ controls how many top-ranked teacher tokens the watermark delta is applied to. \cref{tab:ablation-topk} sweeps $k \in \{50, 100, 200, 300, 500\}$. Post-merge robustness broadly trends upward with $k$ at the cost of PPL; $k\!=\!100$ gives the best PPL/robustness trade-off (lower PPL than the base while matching its post-merge TPR), staying within $1$\,pp of the base standalone TPR, so we move it into the joint search.
Compared to the no-gating baseline of \citet{learnability_wm}, restricting watermark distillation to the top-$k$ logits lowers PPL while leaving standalone TPR essentially unchanged (within $\sim$2\,pp).

\begin{table}[!htbp]
\centering
\caption{\textbf{Top-$k$ ablation:} Vary $k$; hold $\alpha_{\min}\!=\!0.1$, $\delta\!=\!2.0$. Standalone and post-merge TPR@1\% in \%; PPL on \textsc{Llama-3.2-1B}. Best checkpoint values reported.}
\vspace{4pt}
\label{tab:ablation-topk}
\footnotesize
\setlength{\tabcolsep}{4pt}
\begin{tabular}{c rr rr}
\toprule
$k$ & TPR@1\% (st.) & PPL & TPR@1\% ($\alpha\!=\!0.5$) & seq-rep-3 \\
\midrule
50              & 95.7 & 5.58 & 55.7 & 0.097 \\
\textbf{100}    & \textbf{95.3} & \textbf{5.90} & \textbf{60.0} & \textbf{0.094} \\
200 (base)      & 96.0 & 6.25 & 59.7 & 0.083 \\
300             & 97.3 & 6.28 & 65.7 & 0.094 \\
500             & 96.7 & 6.45 & 67.0 & 0.089 \\
\bottomrule
\end{tabular}
\end{table}

\subsection{Watermark Strength ($\delta$)}
\label{sec:ablations:delta}
$\delta$ is the green-list bias added by the teacher. Higher $\delta$ produces a stronger watermark signal but degrades quality. \cref{tab:ablation-delta} sweeps $\delta \in \{1.0, 2.0, 2.3, 2.5, 2.7\}$. $\delta\!=\!1.0$ is too weak (standalone TPR\,$\sim\!55\%$); $\delta\!=\!2.3$ gives the cleanest robustness/quality trade-off, substantially improving post-merge TPR over the base without the PPL inflation we see at $\delta\!=\!2.5$ / $2.7$.

\begin{table}[!htbp]
\centering
\caption{\textbf{Watermark-strength ablation:} Vary $\delta$; hold $k\!=\!200$, $\alpha_{\min}\!=\!0.1$. Best checkpoint values reported.}
\vspace{4pt}
\label{tab:ablation-delta}
\footnotesize
\setlength{\tabcolsep}{4pt}
\begin{tabular}{c rr rr}
\toprule
$\delta$ & TPR@1\% (st.) & PPL & TPR@1\% ($\alpha\!=\!0.5$) & seq-rep-3 \\
\midrule
1.0           & 55.3 & 4.98 & 11.7 & 0.077 \\
2.0 (base)    & 96.0 & 6.25 & 59.7 & 0.083 \\
\textbf{2.3}  & \textbf{98.3} & \textbf{6.69} & \textbf{76.7} & \textbf{0.074} \\
2.5           & 99.0 & 7.11 & 89.3 & 0.086 \\
2.7           & 99.0 & 7.82 & 89.7 & 0.068 \\
\bottomrule
\end{tabular}
\end{table}

\subsection{Merge-Adversary Lower Bound ($\alpha_{\min}$)}
\label{sec:ablations:alpha-min}
The merge adversary samples $\alpha \sim \mathcal{U}(\alpha_{\min}, 1)$ at each step. Lower $\alpha_{\min}$ exposes the student to more aggressive merges. \cref{tab:ablation-alphamin} sweeps $\alpha_{\min} \in \{0.1, 0.3, 0.5\}$. Robustness drops sharply as $\alpha_{\min}$ rises: at $\alpha_{\min}\!=\!0.5$, post-merge TPR@1\% drops by a third vs the base, because the model never trains against $\alpha < 0.5$. $\alpha_{\min}\!=\!0.1$ wins clearly and we keep it for the joint config.

\begin{table}[!htbp]
\centering
\caption{\textbf{Merge-adversary lower-bound ablation:} Vary $\alpha_{\min}$; hold $k\!=\!200$, $\delta\!=\!2.0$. Best checkpoint values reported.}
\vspace{4pt}
\label{tab:ablation-alphamin}
\footnotesize
\setlength{\tabcolsep}{4pt}
\begin{tabular}{c rr rr}
\toprule
$\alpha_{\min}$ & TPR@1\% (st.) & PPL & TPR@1\% ($\alpha\!=\!0.5$) & seq-rep-3 \\
\midrule
\textbf{0.1} (base) & \textbf{96.0} & \textbf{6.25} & \textbf{59.7} & \textbf{0.083} \\
0.3                 & 97.3          & 6.46          & 54.0          & 0.085 \\
0.5                 & 97.7          & 6.37          & 40.3          & 0.083 \\
\bottomrule
\end{tabular}
\end{table}

\subsection{Joint Search}
\label{sec:ablations:joint}
\cref{tab:ablation-joint} validates the per-parameter winners ($k\!=\!100$, $\alpha_{\min}\!=\!0.1$) jointly with a $\delta$ sweep, fixing the other parameters at base. The headline trade-off is $\delta$ vs PPL: $\delta\!=\!2.5$ recovers $88.3\%$ at $\alpha\!=\!0.5$ but pays $+0.90$ PPL over $\delta\!=\!2.0$, while $\delta\!=\!2.3$ recovers $80.3\%$ at $\alpha\!=\!0.5$ for only $+0.26$ PPL. We choose $\boldsymbol{\delta\!=\!2.3,\,k\!=\!100,\,\alpha_{\min}\!=\!0.1}$ as the main-paper config: best PPL/robustness frontier without the seq-rep-3 risk of higher $\delta$. Increasing $k$ from $100$ back to $200$ at the same $\delta$ slightly hurts both robustness ($-3.3$\,pp) and PPL, confirming $k\!=\!100$ as the joint-best.

\begin{table}[!htbp]
\centering
\caption{\textbf{Joint search:} Combine the per-parameter winners ($\alpha_{\min}\!=\!0.1$ throughout) with a $\delta$ sweep; the row in \textbf{bold} is the main-paper configuration.}
\vspace{4pt}
\label{tab:ablation-joint}
\footnotesize
\setlength{\tabcolsep}{4pt}
\begin{tabular}{cc rr rr}
\toprule
$k$ & $\delta$ & TPR@1\% (st.) & PPL & TPR@1\% ($\alpha\!=\!0.5$) & seq-rep-3 \\
\midrule
100 & 2.0          & 96.7 & 6.32 & 60.3 & 0.086 \\
100 & 2.2          & 98.0 & 6.56 & 79.7 & 0.080 \\
\textbf{100} & \textbf{2.3} & \textbf{97.0} & \textbf{6.58} & \textbf{80.3} & \textbf{0.082} \\
200 & 2.3          & 97.7 & 6.82 & 77.0 & 0.078 \\
100 & 2.5          & 97.3 & 7.22 & 88.3 & 0.081 \\
\bottomrule
\end{tabular}
\end{table}

\subsection{Importance of Sampling $\alpha$}
\label{sec:ablations:alpha-sampling}
We test whether sampling $\alpha$ matters at all by replacing the uniform adversary with $\alpha$ \emph{fixed} at $0.3$. \cref{tab:ablation-fixed-alpha} shows the fixed-$\alpha$ run sits below the sampled baseline at \emph{every} evaluated $\alpha$, including $0.3$ itself; robustness collapses away from the training point ($2.7\%$ at $\alpha\!=\!0.1$, $40.0\%$ at $\alpha\!=\!0.5$, $-19.7$\,pp vs.\ baseline) and PPL rises sharply ($9.35$ vs.\ $6.25$) as the student over-fits the single training point. Uniform $\alpha$ sampling outperforms the fixed-$\alpha$ schedule at every $\alpha$ and at standalone PPL, which is why we adopt it as our default.

\begin{table}[!htbp]
\centering
\caption{\textbf{Sampled vs.\ fixed merge-adversary $\alpha$:} Post-merge TPR@1\% across the full $\alpha$ range (Linear merging on C4). Top row: the default merge adversary, which samples $\alpha \sim \mathcal{U}(0.1, 1)$ at every training step (same configuration as $\alpha_{\min}\!=\!0.1$ in \cref{tab:ablation-alphamin}).}
\vspace{4pt}
\label{tab:ablation-fixed-alpha}
\footnotesize
\setlength{\tabcolsep}{4pt}
\begin{tabular}{l ccccc cc}
\toprule
\textbf{Merge-adversary $\alpha$} & \multicolumn{5}{c}{\textbf{Post-merge TPR@1\% at eval $\alpha\!=\!\cdot$}} & \textbf{Standalone TPR@1\%} & \textbf{PPL} \\
\cmidrule(lr){2-6}
 & 0.1 & 0.3 & 0.5 & 0.7 & 0.9 & & \\
\midrule
$\alpha \sim \mathcal{U}(0.1, 1)$ (sampled, default) & 12.7 & 38.7 & 59.7 & 81.0 & 93.7 & 96.0 & 6.25 \\
Fixed $\alpha\!=\!0.3$ (no sampling)                 &  2.7 & 10.0 & 40.0 & 70.0 & 91.3 & 96.7 & 9.35 \\
\bottomrule
\end{tabular}
\end{table}

\section{Existing Assets and Licenses}
\label{app:licenses}

We use the following existing datasets and models. We list the license names as reported by the corresponding dataset or model cards, repositories, or project pages.

\paragraph{Datasets.}
\begin{itemize}
    \item \textsc{Alpaca-GPT4}: Creative Commons Attribution-NonCommercial 4.0 International (CC BY-NC 4.0).
    \item \textsc{OpenWebText}: Creative Commons Zero v1.0 Universal (CC0 1.0) for the dataset packaging; the dataset creators state that they do not own the underlying extracted text.
    \item \textsc{Alpaca-GPT4-DE}: Apache License 2.0.
    \item \textsc{FineWeb-2}: Open Data Commons Attribution License v1.0 (ODC-BY 1.0).
    \item \textsc{MetaMathQA}: MIT License.
    \item \textsc{CodeAlpaca}: Creative Commons Attribution 4.0 International (CC BY 4.0) for the dataset release.
    \item \textsc{NuminaMath-CoT}: Apache License 2.0.
    \item \textsc{Evol-Instruct-DE}: Apache License 2.0.
    \item \textsc{French-Alpaca}: Apache License 2.0.
    \item Lucie Training Dataset: Creative Commons Attribution-NonCommercial-ShareAlike 4.0 International (CC BY-NC-SA 4.0).
    \item \textsc{C4}: Open Data Commons Attribution License v1.0 (ODC-BY 1.0).
    \item \textsc{GSM8K}: MIT License.
    \item \textsc{ARC-Challenge}: Creative Commons Attribution-ShareAlike 4.0 International (CC BY-SA 4.0).
    \item \textsc{MMLU}: MIT License.
    \item \textsc{HellaSwag}: MIT License.
    \item \textsc{MATH}: MIT License.
\end{itemize}

\paragraph{Models.}
\begin{itemize}
    \item \textsc{Llama-3.1-8B-Instruct}: Llama 3.1 Community License.
    \item \textsc{Qwen-2.5-3B-Instruct}: Qwen Research License Agreement.
    \item \textsc{Llama-2-13B}: Llama 2 Community License Agreement.
    \item \textsc{FuseChat-Llama-3.1-8B-Instruct}: Apache License 2.0 as listed by its model card; because it is derived from \textsc{Llama-3.1-8B-Instruct}, we also treat it as subject to the Llama 3.1 Community License.
    \item \textsc{OpenMath2-Llama3.1-8B}: Llama 3.1 Community License.
    \item \textsc{Llama-3.1-Tulu-3-8B}: Llama 3.1 Community License.
\end{itemize}

\section{Broader Impacts}
\label{app:broader-impact}

Our work aims to make provenance signals for open-source LLMs more durable under common downstream modifications. A positive impact is that robust watermarking can help model providers, auditors, and researchers attribute generated text after benign post-training workflows such as finetuning and model merging. This can support accountability and make deployed open-source systems easier to monitor.

The same capability also has limitations and risks. Watermark detection can be misinterpreted if users ignore false-positive rates, key-management assumptions, or the possibility that downstream transformations fall outside the evaluated threat model. More robust watermarks may also strengthen unilateral control over models after release, so deployments should pair detection claims with transparent operating thresholds, appeals or audit procedures where relevant, and clear communication about the detector's scope.

	\endgroup
\fi

\end{document}